%% file: main.tex
\documentclass[]{bytedance_seed}
\usepackage[toc,page,header]{appendix}

\input{math_commands.tex}

\usepackage{makecell}
\usepackage[pagebackref=true,breaklinks=true,letterpaper=true,colorlinks,bookmarks=false,citecolor=blue,linkcolor=blue]{hyperref}
\usepackage{url}
\usepackage{booktabs}
\usepackage{arydshln}
\usepackage{multirow}
\usepackage{graphicx}
\usepackage{subcaption}
\usepackage{enumitem}
\usepackage{xspace}
\usepackage{mathabx}
\usepackage{bbm}
\usepackage{wrapfig}
\usepackage{xcolor}
\usepackage{adjustbox}

\usepackage{geometry} 
\usepackage{tikz} 

\graphicspath{{figures/}}

\usepackage[capitalize]{cleveref}
\crefname{section}{Section}{Secs.}
\Crefname{section}{Section}{Sections}
\Crefname{table}{Table}{Tables}
\crefname{table}{Tab.}{Tabs.}

\makeatletter
\DeclareRobustCommand\onedot{\futurelet\@let@token\@onedot}
\def\@onedot{\ifx\@let@token.\else.\null\fi\xspace}

\def\eg{\textit{e.g}\onedot} 
\def\ie{\textit{i.e}\onedot} 
 
\def\etc{\textit{etc}\onedot}

\makeatother

\definecolor{mygray}{gray}{0.4}

\newcommand{\modelname}{Seaweed-7B}

\title{Seaweed-7B: Cost-Effective Training of \\
Video Generation Foundation Model}

\author{ByteDance Seed}

\affiliation{ByteDance}

\input{sections/0_abstract}

\date{March 02, 2025}

\checkdata[Project Page]{\url{https://seaweed.video/}}

\begin{document}

\maketitle

\input{sections/1_introduction}

\input{sections/2_data}
\input{sections/3_method}
\input{sections/4_evaluation}

\input{sections/5_applications}
\input{sections/6_related_work}

\input{sections/7_conclusion}


{
\small
\bibliographystyle{plainnat}
\bibliography{main}
}
\clearpage





\end{document}

%% file: math_commands.tex
\usepackage{amsmath,amsfonts,bm}

\def\eqref#1{equation~\ref{#1}}

\def\1{\mathbbm{1}}

\DeclareMathAlphabet{\mathsfit}{\encodingdefault}{\sfdefault}{m}{sl}
\SetMathAlphabet{\mathsfit}{bold}{\encodingdefault}{\sfdefault}{bx}{n}



%% file: sections/0_abstract.tex
\abstract{
This technical report presents a cost-efficient strategy for training a video generation foundation model.
We present a mid-sized research model with approximately 7 billion parameters (7B) called Seaweed-7B trained from scratch using 665,000 H100 GPU hours. Despite being trained with moderate computational resources, Seaweed-7B demonstrates highly competitive performance compared to contemporary video generation models of much larger size. Design choices are especially crucial in a resource-constrained setting. This technical report highlights the key design decisions that enhance the performance of the medium-sized diffusion model. Empirically, we make two observations: (1) Seaweed-7B achieves performance comparable to, or even surpasses, larger models trained on substantially greater GPU resources, and (2) our model, which exhibits strong generalization ability, can be effectively adapted across a wide range of downstream applications either by lightweight fine-tuning or continue training.
}

%% file: sections/1_introduction.tex
\section{Introduction}
\label{sec:intro}

Foundation models serve as the cornerstone of modern machine learning. These models typically contain a massive number of parameters and are trained on vast amounts of data, allowing them to demonstrate strong generalization capabilities and adapt to a diverse range of downstream tasks. Examples include large language models (LLMs) for natural language processing~\cite{brown2020language,chowdhery2023palm}, vision language models for image/video understanding~\cite{radford2021learning,alayrac2022flamingo}, and audio foundation models for speech synthesis and recognition~\citep{borsos2023audiolm,radford2023robust}.
This paper focuses on the foundation model for video generation, a compelling research area driven by the central role of video as a dominant medium in digital entertainment, communication, and real-world simulation.
The video generation model plays a pivotal role, as advancements in this foundation can broadly enhance performance across a range of downstream video applications such as image animation~\cite{kondratyuk2023videopoet,chen2025livephoto}, video editing~\cite{firefly}, and video storytelling~\cite{guo2025long,xiao2025videoauteur}.

Video generation models have seen rapid advancements in the past few years. Recent reports present various methods for training video generation models from scratch, such as MovieGen~\cite{polyak2025moviegencastmedia}, Cosmos~\citep{agarwal2025cosmos}, and Wan-2.1~\citep{wan2.1}, among many others. These approaches exhibit a consistent pattern, utilizing diffusion transformers (DiT)~\citep{peebles2023scalable, esser2024scaling} and adhering to the trend of scaling the model size, along with the GPU resources, to improve performance. Scaling up DiT models holds promise, but its training demands a massive GPU cost.
For example, MovieGen uses 6,000+ NVIDIA H100 GPUs. Such demands can significantly impede innovation in video generation models.

Beyond the high training costs, inference in video generation remains exceptionally expensive which is often orders of magnitude more than language, image, or audio generation. For many applications, such as those in social media like Instagram and YouTube Shorts, inference may be constrained by GPU memory and the high serving costs. As a result, the substantial training and inference expenses tend to favor small to medium-sized models, which offer better cost efficiency for both training and inference.

Fortunately, the language model community has discovered that small to medium-sized models can match or even surpass large language models (LLMs) through architectural improvements and optimized training strategies~\citep{jiang2023mistral7b,liu2024deepseek}. For instance, Mistral 7B outperforms Llama2 13B across benchmarks~\citep{jiang2023mistral7b}. DeepSeek v3~\cite{liu2024deepseek} demonstrates that a 37B-parameter activation model can surpass 72B and 420B dense models, requiring only a fraction of GPU resources. This efficiency is achieved through key designs such as enhanced Mixture-of-Experts (MoE), Multi-Token Prediction (MTP), and the use of high-quality training data.

In video generation, however, few studies have investigated similar scaling efficiencies\footnote{Our work focuses on training a video generation model from scratch, which distinguishes it from model compression techniques such as quantization and distillation, which build on a pre-trained diffusion transformer model.}. Although earlier works have explored training small models~\citep{lin2024open,opensora} with minimal GPU resources, their impact remains limited due to a significant quality gap between their generated videos and those by contemporary state-of-the-art models.

This technical report discusses a cost-efficient strategy for training a video generation foundation model. We choose to train a moderately sized model with FLOPs optimized for deployment on a single GPU, namely \textit{\modelname{}} (short for Seed Video),
which consists of a DiT with approximately 7 billion parameters. We train the model from scratch using 665,000 H100 GPU hours, equivalent to 27.7 days of training on 1,000 H100 GPUs. Fortuitously, we have trained versions of the model with similar model sizes and GPU resources. This allows us to carry out meaningful comparisons of their design differences. Our findings indicate the critical impact of design choices in this resource-constrained setting, particularly in data curation, model design, and training strategy and optimization.

To validate the performance of \modelname{} as a foundational model for video generation, we conduct experiments evaluating two hallmark capabilities of foundation models as discussed in~\citep{bommasani2021opportunities}: generic generation capability and downstream task generalization. First, we evaluate two primary tasks, \ie, text-to-video and image-to-video generation, to assess generation quality in terms of fidelity, aesthetics, motion quality, prompt alignment, and inference efficiency. Our results show that \modelname{} matches or even surpasses some significantly larger models trained with greater computational resources, showcasing its highly competitive performance. Second, we perform a qualitative analysis of adapting \modelname{} across a variety of video generation tasks. The results demonstrate that \modelname{} can be effectively applied to a broad range of downstream applications, either by lightweight fine-tuning or continue training (see \Cref{sec:applications}).

Our experimental results suggest that the potential of a medium-sized DiT model, such as those with 7 billion parameters, remains largely underexplored. Given their cost-efficiency advantages in both training and inference, we hope future research will continue to optimize medium-sized models.

The structure of this paper is as follows. Since previous works have extensively detailed video generation model designs, this paper focuses on key design choices that complement or enhance existing findings in the literature. In summary, our contributions are as follows:

\begin{itemize}
    \setlength{\itemsep}{0pt}
    \item Variational Autoencoder (VAE) designs that achieve state-of-the-art reconstruction quality. We share key insights on balancing the trade-offs between compression efficiency, reconstruction fidelity, and generation quality.
    \item Insights and lessons learned from training Diffusion Transformers (DiT), including cost-effective training strategies and architectural considerations.
    \item Empirical evidence demonstrating the competitive performance of a medium-sized model across multiple video generation tasks.
\end{itemize}

%% file: sections/2_data.tex
\section{Data}
In a constrained computing setting, data quality and diversity take precedence over quantity. To collect high-quality video data, we describe a scalable infrastructure for large-scale data processing, and various data processors for effectively scanning high-quality video data. Using our data pipeline, we collect data at a scale of $\Theta(100M)$ clips, each with an average duration of about 8 seconds.

\subsection{Data Processing}
\label{sec:data_process}
Our raw video data pool originates from diverse sources.
To transform this into high-quality training data, we employ a comprehensive data curation pipeline, which includes but is not limited to, temporal splitting, spatial cropping, quality filtering, multi-aspect data balancing, video deduplication, and video captioning. 

\paragraph{Temporal splitting.} Raw videos are first split into single-shot clips via our internal splitter, which detects shot boundaries via HSV 3D color histogram feature similarity of adjacent sampled frames. This lightweight method performs comparably to pyscenedetect~\cite{pyscenedetect} and excels in fade/dissolve transitions. In addition, we merge single-shot clips with the ImageBind features~\cite{girdhar2023imagebind} to form multi-shot sequences for long video generation~\cite{xiao2025videoauteur,guo2025long}.

\paragraph{Spatial cropping.} We utilize the crop-detect filter in FFmpeg~\cite{ffmpeg} for black border removal and develop frame-level models for text, logo, watermark, and special effects detection. A bounding box aggregation algorithm, considering confidence and IoU, is used to aggregate these frame-level detections into video-level bounding boxes, ensuring high detection accuracy. This algorithm assumes that graphical overlays have fixed sizes and locations and appear consistently across consecutive frames. Finally, optimal cropping regions are selected using a heuristic method. If no suitable cropping region can be found due to the size, location, or aspect ratio of detected unwanted regions, the clips will be discarded.

\paragraph{Quality filtering.} Clips undergo a multi-stage sequential filtering process. 1) Attribute filtering: retain clips with $5$--$60$ seconds, short-side $\geq256$ px, aspect ratio within $[1/3, 3/1]$. 2) Visual quality evaluation: use our specialized visual quality model to estimate aesthetics and clarity scores, followed by clip removal using manually selected thresholds tailored to each data source and video type. 3) Spatial-temporal motion filtering: eliminate static clips and undesirable movements by an improved motion amplitude aggregation algorithm based on motion vectors~\cite{ffmpeg}. This strategy achieves comparable performance to optical flow~\cite{teed2020raft} for our task while offering higher efficiency. 4) Camera shake and playback speed detection: utilize a high FPS detector to identify unstable camera movements or inconsistent playback speeds. 5) Safety screening: remove harmful content including violent scenes, pornography, and nudity. 6) Artifact detection: employ several classifiers to detect non-natural effects such as slide transitions, speed-up, jitter, oversaturation, \etc. This pipeline reduced the invalid clip rate from $42\%$ to $2.9\%$, as confirmed by manual evaluation.




\paragraph{Multi-aspect data balancing and video deduplication.}
Raw video content exhibits a long-tailed distribution across subjects, scenes, and actions. To address this imbalance, we cluster both visual and semantic features extracted from the videos into over 10,000 groups. The visual features are obtained from a CLIP-like model~\cite{radford2021learning}, while the semantic features are derived from LLM-generated labels based on video captions. Clustering based on visual and semantic features enables effective detection and removal of duplicate content within the training data. Subsequently, we downsample head categories to smooth the distribution for training, while preserving both visual and semantic diversity.
This strategy has been proven to be an effective and cost-efficient approach for model training.

\paragraph{Simulation data.} 
We also use computer-generated (synthetic) videos to augment the long-tailed distribution of our training data. Our goal is to enhance video generation in terms of 3D consistency in camera movements, and body integrity during complex human motion~\cite{zhao2025synthetic}. To do so, we develop a synthetic video generation pipeline that simulates 3D scenes with fine-grained control, leveraging high-quality object, human, and motion assets within graphics engines~\cite{Blender, UE5}. For more details, see~\cite{zhao2025synthetic}. By planning both visual and control parameters, we render a few million synthetic videos and mix them with real-world videos during training.

\begin{figure}[h!]
  \centering
  \includegraphics[width=.98\linewidth]{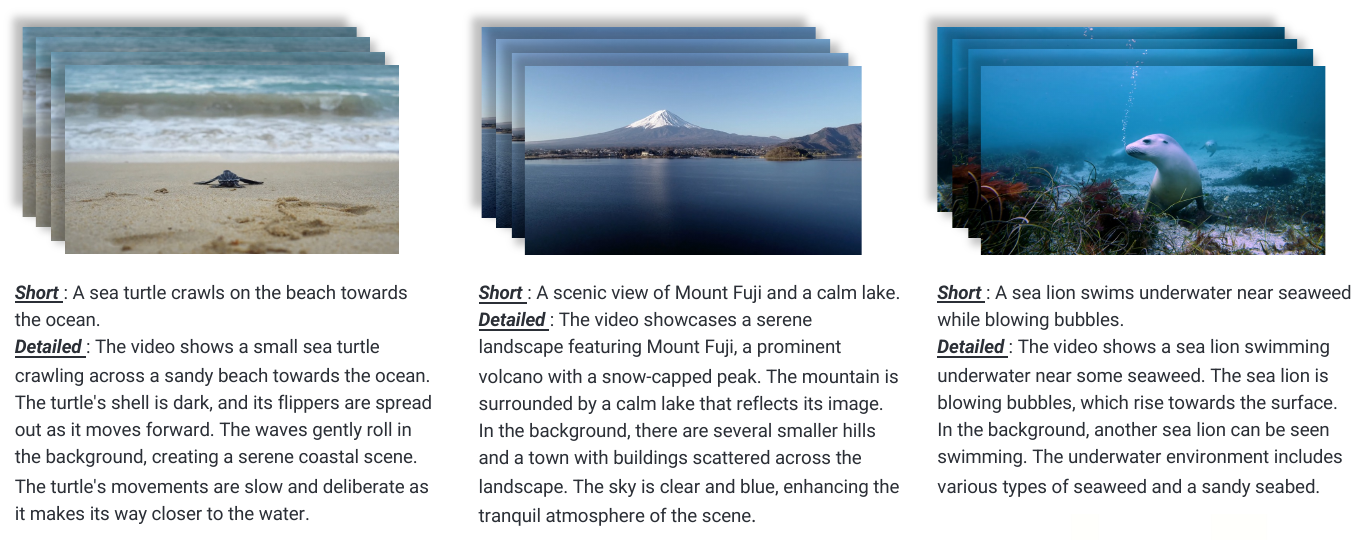}
   \vspace{-2mm}
  \caption{Short and detailed captions are generated by our video captioning model. The short captions provide action-centric summaries of the videos, while the detailed captions offer rich descriptions of the scenes, including attributes, objects, and environments.}
  \vspace{-3mm}
  \label{fig:caption}
\end{figure}

\paragraph{Video captioning.} 
Video captions are essential for enhancing prompt-following capabilities. To improve this ability, we train a dedicated video captioning model. A key strategy involves generating both short and detailed captions for the entire training set of video clips. The short captions provide action-centric summaries of the input videos, while the detailed captions include richer visual descriptions of scenes, objects, attributes, and more. \Cref{fig:caption} shows some generated caption samples. Our video captioning model consists of a pre-trained CLIP~\citep{radford2021learning} vision encoder and an LLM for captioning.

To enhance captioning accuracy, we uniformly sample 32 frames from each video as input for captioning. To balance efficiency with accuracy, we apply AnyRes~\cite{liu2024improved} only to 8 frames, while the remaining 24 frames are center-cropped. The model trained with these additional center-cropped frames produces captions with reduced hallucinations. Furthermore, the video timestamps used for captioning are aligned with those during DiT training.

We observe that employing a larger 72B LLM reduces hallucinations in our captioning task. However, generating video captions for millions of videos using the 72B model entails significantly higher computational costs. Therefore, we utilize the 72B LLM as a teacher model to distill knowledge into a more computationally efficient 7B student model, thus improving captioning accuracy without incurring additional inference overhead.  
Moreover, our findings indicate that initially generating a ``detailed'' caption, followed by the derivation of a ``short'' caption, analogous to a chain-of-thought process, further enhances the accuracy of the short caption. On our test set, this inference strategy increases the accuracy from 84. 81\% to 90. 84\%.

\paragraph{System Prompt.} In addition to the captions discussed above, we incorporate specific subject-oriented descriptions to further characterize each video. These include video types (\eg, PGC, UGC), camera position (\eg, medium shot, close-up shot), camera angles (\eg, low-angle shot, eye-level shot), camera movement (\eg, zoom-in, pan left, arc shot) and visual styles (\eg, vintage, cinematic, cartoon). Some labels, such as video types, can be directly extracted from metadata. For others, while it is possible to generate labels using the same captioning models, we find it more accurate to apply classifiers trained specifically for these attributes and retain only high-confidence predictions from these models. Since these labels serve a different purpose than captions, we refer to them as \textit{system prompts}. During training, system prompts are randomly sampled and appended to the video captions. This setup enables the use of system prompts during video generation inference.

\subsection{High-Throughput Pipeline} 

\begin{figure}[h]
  \centering
  \includegraphics[width=.9\linewidth]{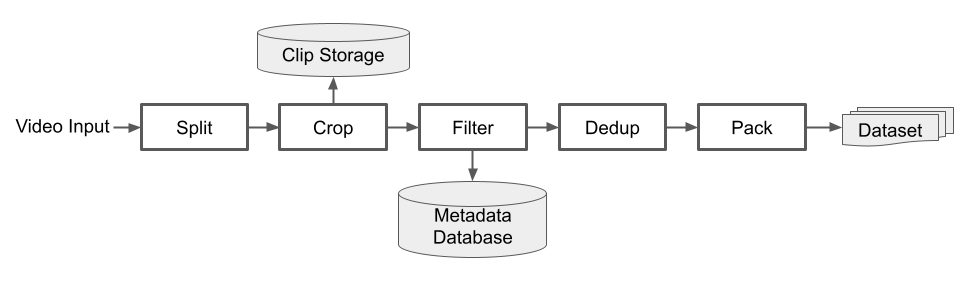}
  \vspace{-5mm}
  \caption{Video Data Processing Pipeline Overview.}
  \label{fig:video_pipeline}
   \vspace{-5mm}
\end{figure}

Data forms the foundation of the video generation model. To efficiently process and retrieve video data at scale, we have developed a high-throughput and flexible video curation pipeline.
This pipeline is designed to manage video encoding and decoding, perform temporal segmentation and spatial cropping, and apply all video quality filtering operations outlined in \Cref{sec:data_process}. Using this system, we identify high-aesthetic, high-clarity, and dynamic-rich clips from a vast repository of video data. With this infrastructure, we are able to process over 500,000 hours of video data daily. Given that the quality is more than sufficient for training, the focus is on how to effectively mine high-quality clips using the various data processors outlined in \Cref{sec:data_process}.

Compared to other media types, video data is significantly larger and requires substantially more computational resources. To optimize throughput when processing video clips, we use two modern frameworks: BMF (Babit Multimedia Framework)~\citep{bmf} and Ray~\citep{ray}. 
BMF is a customizable multimedia processing framework, which provides a simple and easy-to-use cross-language interface and can dynamically expand, manage, and reuse the atomic capabilities of video processing in a modular manner. Ray is an open-source compute engine for scaling AI and Python applications. It provides us with an easy way to run distributed jobs easily in a large cluster with both CPU and GPU resources.

%% file: sections/3_method.tex
\section{Design and Discussions}

In this section, we introduce the components of Seaweed that consist of a variational autoencoder (VAE) and a latent diffusion transformer (DiT).
\Cref{subsec:vae} presents the key designs of VAE, including the architecture, the effect of compression ratio, training strategy, and stability. 
\Cref{subsec:dit} further introduces DiT comprehensively, followed by Sec.~\ref{subsec:training} which shares the training strategy for multi-stage, multi-task learning. 
\Cref{subsec:optim} shows the optimization for both training and inference, for cost-effective use of GPU compute.


\subsection{Variational Autoencoder}\label{subsec:vae}

\begin{figure}[h!]
  \centering
  \includegraphics[width=.8\linewidth]{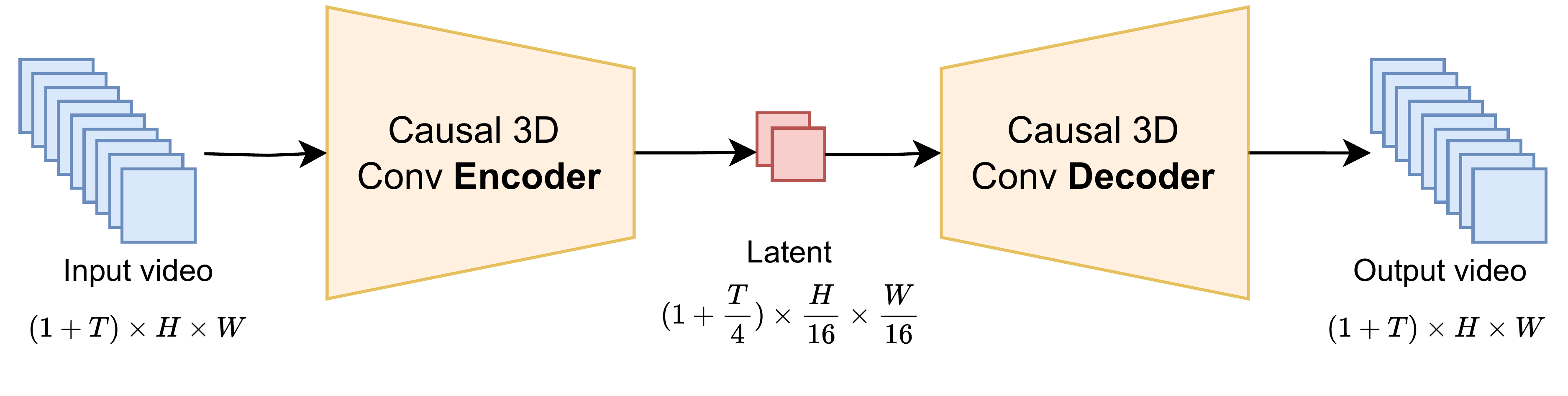}
  \caption{Overview of VAE architecture.}
  \label{fig:vae-archi}
\end{figure}

Variational autoencoders (VAEs)~\citep{kingma2013auto} are commonly used in modern large-scale image and video generation models~\cite{rombach2022high,yu2023language} for efficient training and inference.
As shown in \Cref{fig:vae-archi}, a VAE consists of an encoder that compresses raw pixel data into a compact latent space and a decoder that reconstructs the original input pixels from these latent features.
An ideal VAE should achieve a high compression ratio while maintaining high reconstruction quality. The VAE is an important component because it sets the upper bound for the realism and the fidelity of generated content, and its latent distribution affects the convergence speed of the subsequent generative model.


Following MAGVIT~\citep{yu2023language}, we use a temporally causal convolutional architecture for both the encoder and decoder, which enables image and video compression across space and time in the joint latent. Specifically, the model encodes videos and images from the RGB pixel space of shape $(T'+1, H', W', 3)$ to a continuous latent space of shape $(T+1, H, W, C)$, where $(t, h, w, c)$ denotes time, height, width, and channel dimensions. The downsample ratios are calculated from $d_t = \frac{T'}{T}$, $d_h = \frac{H'}{H}$, and $d_w = \frac{W'}{W}$, where the first frame is always encoded as a dedicated latent representation. The overall compression ratio is given by
\begin{equation}
    r = \frac{C \times T \times H \times W}{3 \times T' \times H' \times W'} = \frac{C}{3 \times d_t \times d_h \times d_w}.
    \label{equa:compression_ratio}
\end{equation}
For simplicity, we denote the VAE model with $(d_t, d_h, d_w, C)=(4,8,8,16)$ as $48\times$ Seaweed VAE and the one with $(d_t, d_h, d_w, C)=(4,16,16,48)$ as $64\times$ Seaweed VAE, respectively according to the compression ratio.

This causal design offers two advantages for video generation. First, it unifies image and video encoding, making the first-frame conditioned image-to-video generation task natural. Second, it eliminates flickering at the boundaries between two inferred clips and allows for encoding and decoding arbitrarily long videos without artificial stitching. In the remainder of this subsection, we share our key observations in VAE design.


\textbf{Compression ratio determines reconstruction quality while downsample ratio affects convergence speed.}
As shown in \cref{equa:compression_ratio}, the compression ratio is determined by the downsample ratios and the latent channels.
We find that the reconstruction quality of VAE mainly depends on the compression ratio. Although VAEs with the same compression ratio converge to similar results, their convergence speeds vary with the downsample ratios. Smaller downsample ratios generally lead to faster convergence.

\begin{figure}[h]
  \centering
  \begin{subfigure}[b]{0.3\textwidth}
    \centering
    \includegraphics[width=\textwidth]{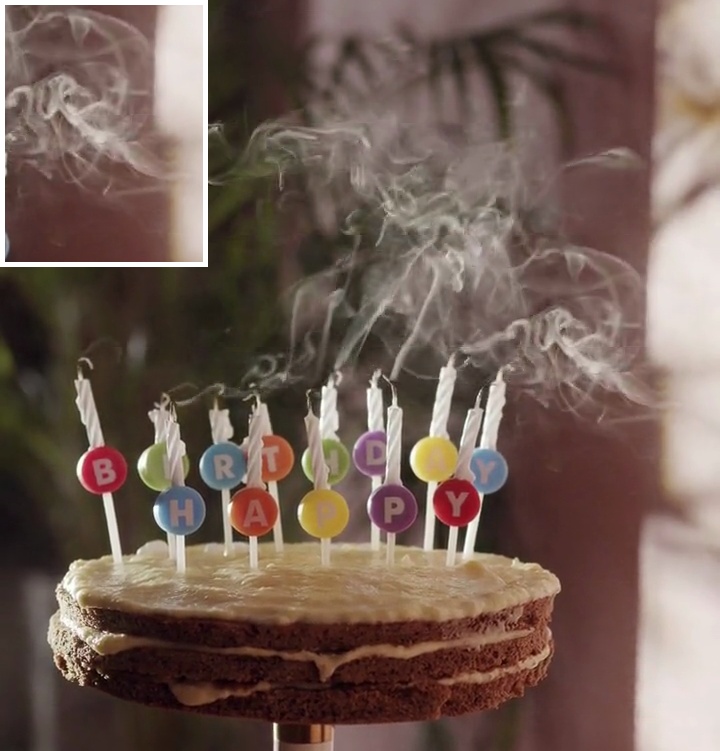}
    \caption{Original Video}
    \label{fig:vae_vis_original_2}
  \end{subfigure}
    \hfill
  \begin{subfigure}[b]{0.3\textwidth}
    \centering
    \includegraphics[width=\textwidth]{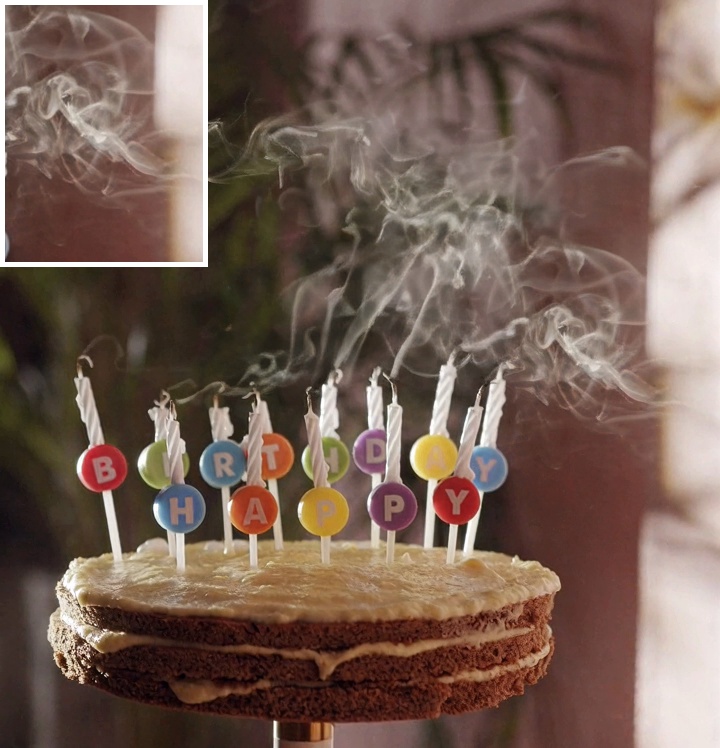}
    \caption{$48\times$ compression Seaweed VAE}
    \label{fig:vae_vis_seaweed1_2}
  \end{subfigure}
  \hfill
  \begin{subfigure}[b]{0.3\textwidth}
    \centering
    \includegraphics[width=\textwidth]{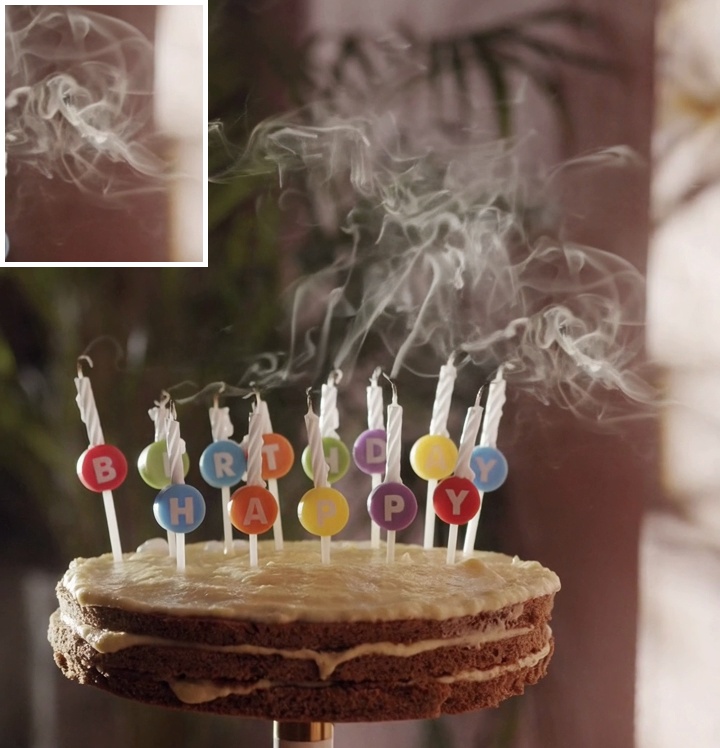}
    \caption{$64\times$ compression Seaweed VAE}
    \label{fig:vae_vis_seaweed2_2}
  \end{subfigure}
  \vspace{-2mm}
  \caption{VAE visualization comparison at 25 fps, with a resolution of 720$\times$720.}
   \vspace{-3mm}
  \label{fig:vae_vis_sample_2}
\end{figure}

\begin{figure}[h]
  \centering
  \begin{subfigure}[b]{0.3\textwidth}
    \centering
    \includegraphics[width=\textwidth]{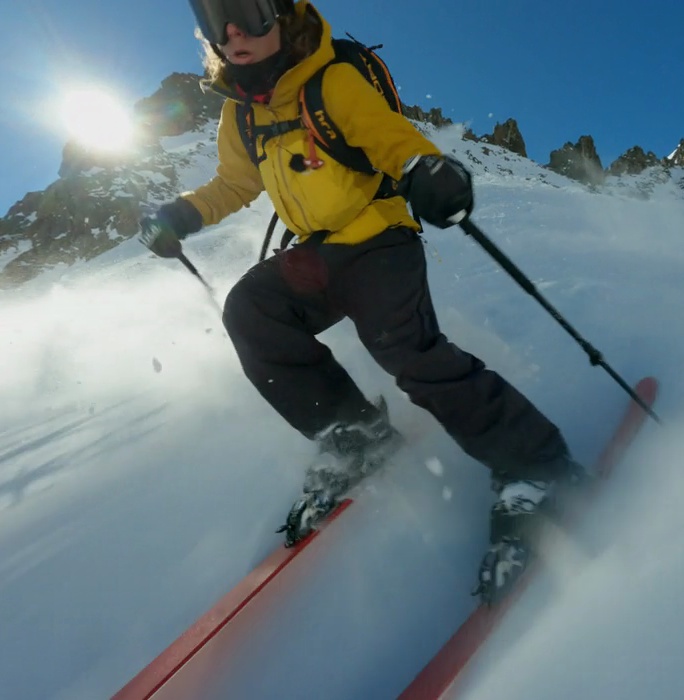}
    \caption{Original Video}
    \label{fig:vae_vis_original_3}
  \end{subfigure}
    \hfill
  \begin{subfigure}[b]{0.3\textwidth}
    \centering
    \includegraphics[width=\textwidth]{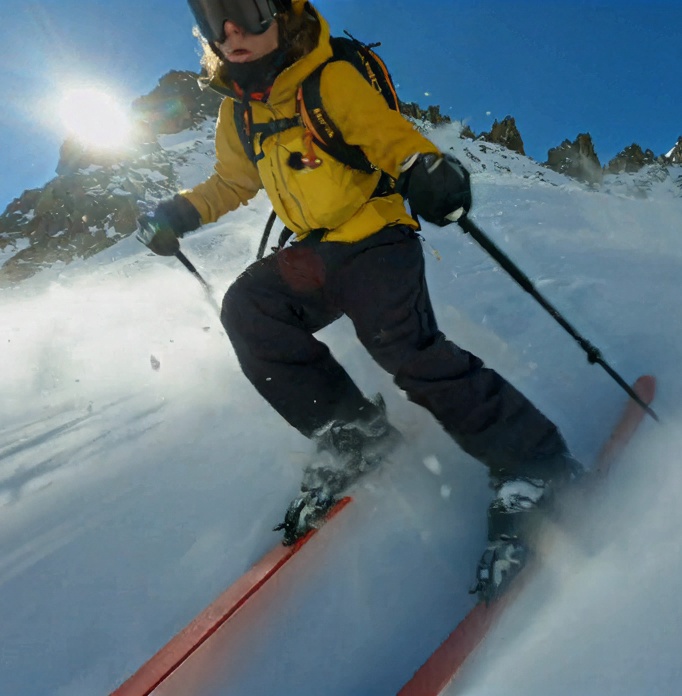}
    \caption{$48\times$ Seaweed VAE}
    \label{fig:vae_vis_seaweed1_3}
  \end{subfigure}
  \hfill
  \begin{subfigure}[b]{0.3\textwidth}
    \centering
    \includegraphics[width=\textwidth]{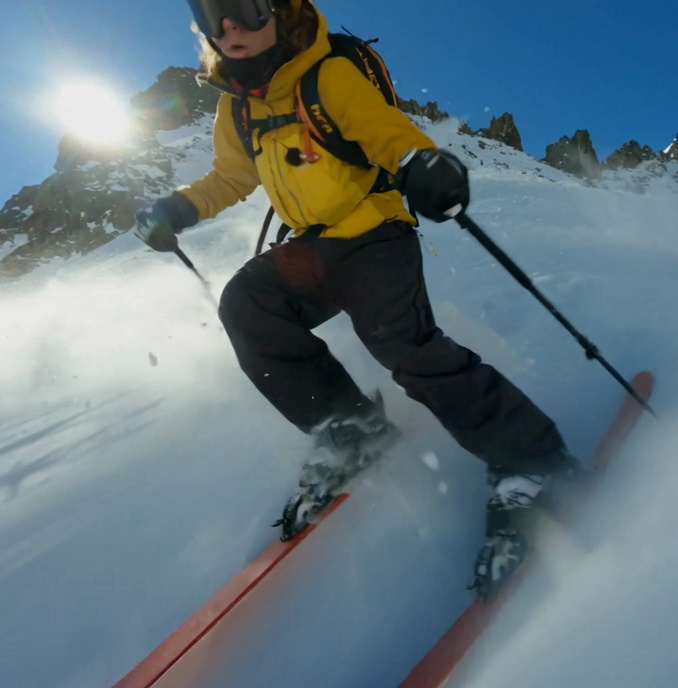}
    \caption{$64\times$ Seaweed VAE}
    \label{fig:vae_vis_seaweed2_3}
  \end{subfigure}
  \vspace{-2mm}
  \caption{VAE visualization comparison at 24 fps, with a resolution of 684$\times$684.}
  \label{fig:vae_vis_sample_3}
\end{figure}

Due to VAE's function of connecting latent space and pixel space, the reconstruction quality of VAE itself reflects the loss of information by compression and significantly affect the fidelity of generation tasks. As shown in \Cref{fig:vae_vis_sample_2,fig:vae_vis_sample_3}, our VAE model effectively reconstructs fine textures and high-dynamic videos, which may be a primary factor contributing to the high realism and vivid motion in our generated videos.

\begin{figure}[htbp]
    \centering
    \begin{minipage}{0.48\linewidth}
        \centering
        \includegraphics[width=\linewidth]{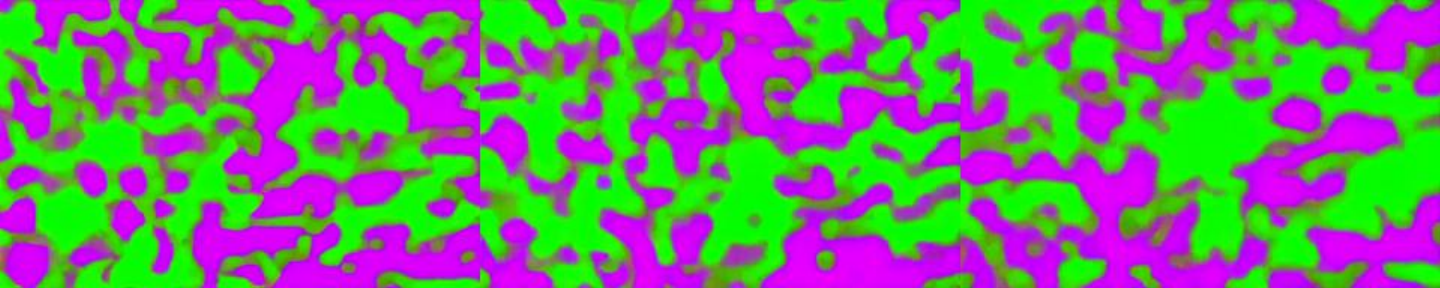}
        \vspace{-.8cm}
        \caption*{$48\times$ VAE at \textbf{30k} steps}
    \end{minipage}
    \begin{minipage}{0.48\linewidth}
        \centering
        \includegraphics[width=\linewidth]{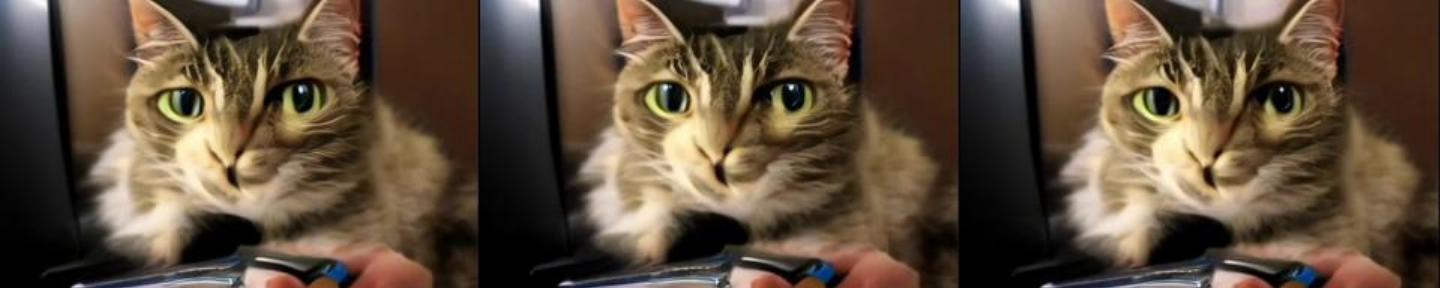}
        \vspace{-.8cm}
        \caption*{$64\times$ VAE at \textbf{30k} steps}
    \end{minipage}
        
    \begin{minipage}{0.48\linewidth}
        \centering
        \includegraphics[width=\linewidth]{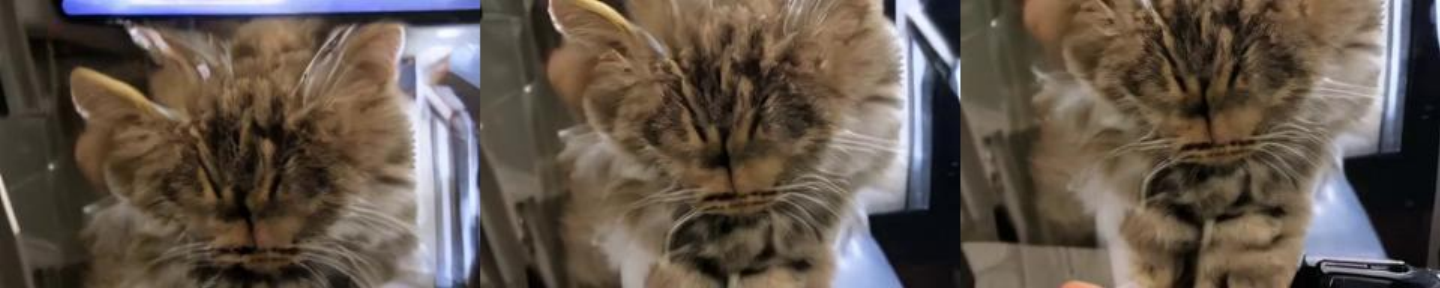}
        \vspace{-.8cm}
        \caption*{$48\times$ VAE, \textbf{45k} steps}
    \end{minipage}
    \begin{minipage}{0.48\linewidth}
        \centering
        \includegraphics[width=\linewidth]{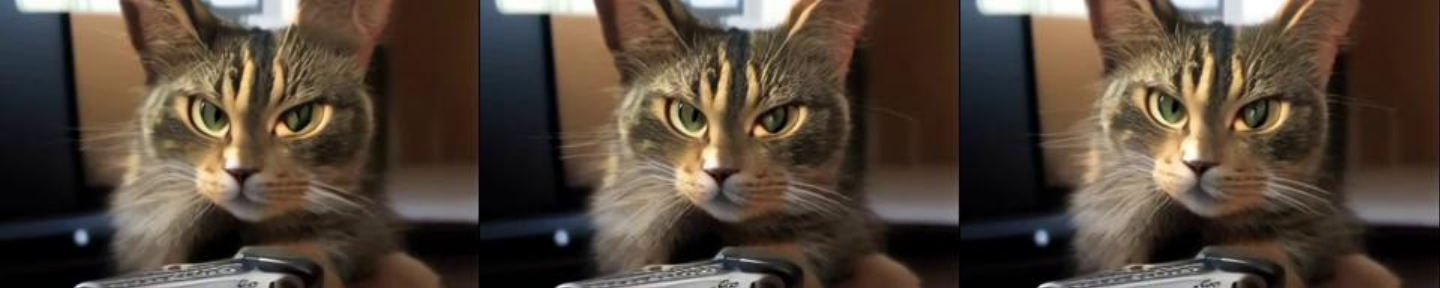}
        \vspace{-.8cm}
        \caption*{$64\times$ VAE at \textbf{45k} steps}
    \end{minipage}
        
    \begin{minipage}{0.48\linewidth}
        \centering
        \includegraphics[width=\linewidth]{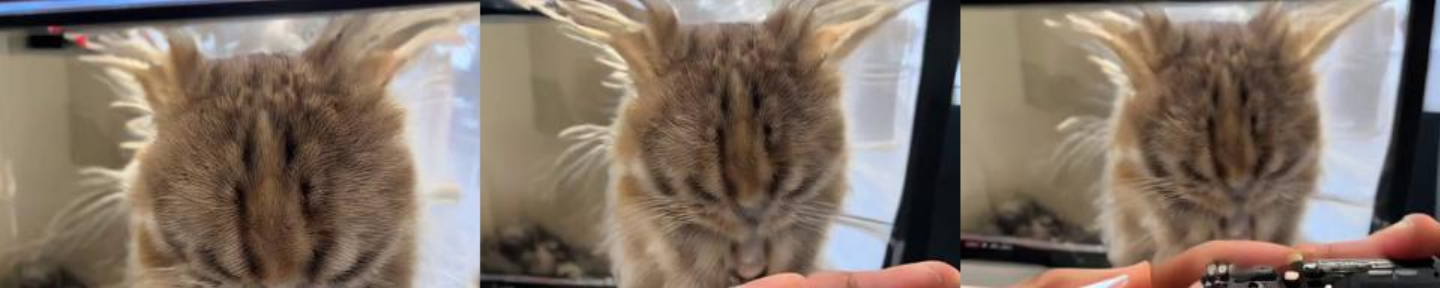}
        \vspace{-.8cm}
        \caption*{$48\times$ VAE at \textbf{60k} steps}
    \end{minipage}
    \begin{minipage}{0.48\linewidth}
        \centering
        \includegraphics[width=\linewidth]{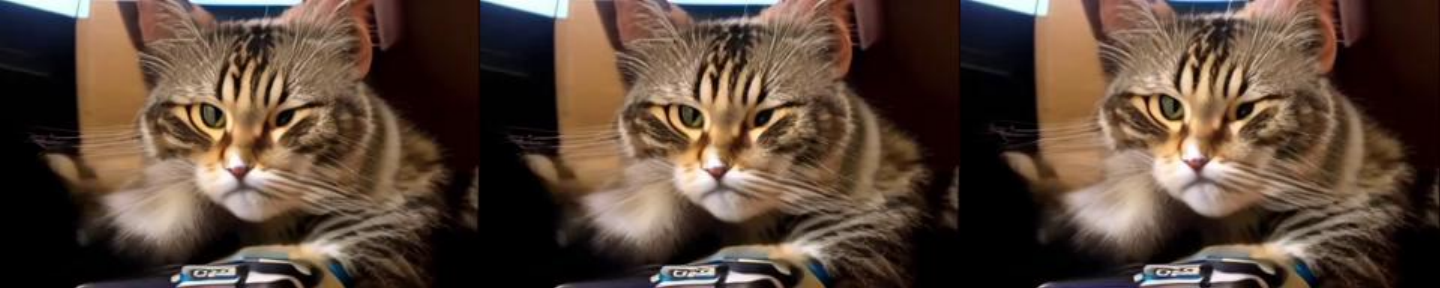}
        \vspace{-.8cm}
        \caption*{$64\times$ VAE at \textbf{60k} steps}
    \end{minipage}
    
    \caption{
        DiT generation results: keyframes from $73 \times 192 \times 320$ videos with the prompt \textit{"Zoom in, cat watching TV with a remote in hand, highly detailed"} are shown. Under the same compute, $64\times$ Seaweed VAE converges faster with a higher compression ratio, without using a DiT patchify.
        \textbf{Left}: $48\times$ Seaweed VAE $(d_t, d_h, d_w) = (4, 16, 16)$ with DiT patch size $(p_t, p_h, p_w) = (1, 2, 2)$. 
        \textbf{Right}: $64\times$ Seaweed VAE $(d_t, d_h, d_w) = (4, 32, 32)$ with DiT patch size $(p_t, p_h, p_w) = (1, 1, 1)$.
    }
    \label{fig:dit-vae-convergence}
    \vspace{-6mm}
    \end{figure}

\textbf{Compressing sequence within VAE outperforms DiT patchification.}
Patchification is commonly used in DiTs~\citep{peebles2023scalable} to merge neighboring tokens, aiming at reducing sequence length and thereby lowering the attention computational cost. For instance, applying patchify with a factor of $(1, 2, 2)$ merges $1\times$ in the temporal dimension and $2\times2$ in both height and width, which reduces the sequence length by a factor of 4.

While VAEs reduce sequence length, they struggle to maintain reconstruction quality under high compression. However, our results show that compressing information through a VAE, while difficult, outperforms patchification in DiT models.
Specifically, we compare two VAE models. The first is a $64\times$ VAE with $(d_t, d_h, d_w) = (4, 16, 16)$ using a $(1, 1, 1)$ patch size; the second is a $48\times$ VAE with $(d_t, d_h, d_w) = (4, 32, 32)$, followed by patchification with a $(1, 2, 2)$ patch size. Notably, both VAEs result in the same sequence length and attention computation cost. The key difference lies in the compression stage: the $64\times$ VAE compresses early within the VAE, while the $48\times$ VAE applies compression later within the DiT input. 

We find that compressing the sequence within the VAE significantly outperforms compression via patchification. $64\times$ VAE not only converges much faster, as shown in \Cref{fig:dit-vae-convergence}, but also converges to a better stationary point. Notably, although the $64\times$ VAE has a higher spatial compression ratio, we do not observe any noticeable visual artifacts in higher-resolution video generation (\eg, 720p).

\textbf{Mixed-resolution training leads to better generalization for high-resolution and long-duration reconstruction.} The extrapolation ability of Causal Conv3D VAEs is limited. Since VAEs are often trained on lower-resolution videos for faster convergence, their performance tends to degrade when decoding high-resolution content. This degradation is partly due to padding operations in both temporal and spatial dimensions, which introduce discrepancies between training and inference. To address this, we propose a mixed-resolution training approach.

\begin{wraptable}{r}{0.45\textwidth}
 \small
  \vspace{-5mm}
  \centering
  \begin{tabular}{c|c|c}
    \toprule
    \textbf{Source} & \textbf{Iteration} & \textbf{Resolution} \\
    \hline
    \multirow{3}{*}{\textbf{Image}} & \multirow{3}{*}{500K} 
    & 720$\times$720            \\
    & & 480$\times$480             \\
    & & 256$\times$256            \\
    \hline
    \multirow{4}{*}{\textbf{Video}} & \multirow{4}{*}{800K} 
    & 17$\times$256$\times$256  \\
    & & 9$\times$480$\times$480  \\
    & & 33$\times$256$\times$256 \\
    & & 113$\times$144$\times$144  \\
    \bottomrule
  \end{tabular}
  \caption{VAE training stages for images and videos. Both stages use mixed-resolution data.}
  \label{table:resolutions}
  \vspace{-15mm}
\end{wraptable}


We train VAE using a diverse set of images and videos with varying resolutions, durations, and frame rates. It is first trained on only images for faster convergence and then on videos. Including high-resolution images and videos in the training improves high-resolution reconstruction quality. The mixed-size training schedule is listed in \Cref{table:resolutions}. We only use 12 FPS and 24 FPS, which are used by the subsequent generative model.

\begin{figure}[h!]
  \centering
  \includegraphics[width=.95\linewidth]{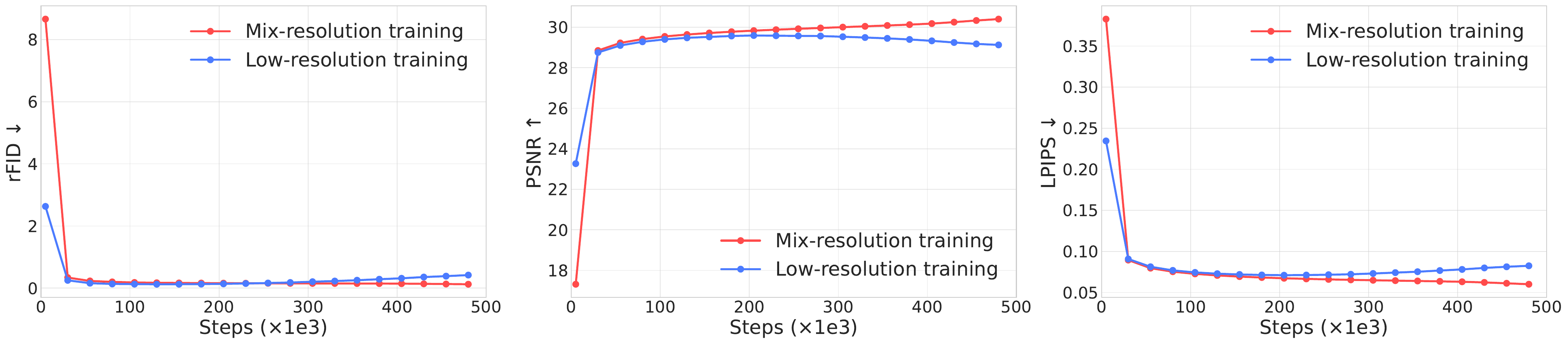}
  \vspace{-3mm}
  \caption{
  Validation metric curves on high-resolution image reconstruction ($512\times512$) show the effectiveness of mix-resolution VAE training.}
  \vspace{-3mm}
  \label{fig:vae_mix_training}
\end{figure}

We evaluate how well the image VAEs reconstruct high-resolution images after being trained in two ways: using only low-resolution images ($256\times256$ pixels) and using a mix of low and high-resolution images ($256\times256$ and $512\times512$ pixels). As shown in \Cref{fig:vae_mix_training}, the VAE trained solely on low-resolution images converges faster initially, but its performance plateaus and even declines towards the end of training. In contrast, the VAE trained on the mixed-resolution dataset consistently improves its ability to reconstruct high-resolution images throughout the training process.


\textbf{Recipes for improving VAE training stability.}
Our VAE is trained using both the reconstruction losses (\ie L1, LPIPS~\cite{zhang2018unreasonableeffectivenessdeepfeatures}) and an adversarial loss. The stability of adversarial training is key for large-scale VAE training. We share effective practices to improve training stability. First, we find using both an image discriminator and a video discriminator yields better results than using either one alone. For the discriminator, PatchGAN \citep{isola2018imagetoimagetranslationconditionaladversarial} is more effective than StyleGAN \citep{karras2020analyzingimprovingimagequality} and UNet \citep{schönfeld2021unetbaseddiscriminatorgenerative} discriminators. However, PatchGAN with BatchNorm~\citep{ioffe2015batchnormalizationacceleratingdeep} can be too strong for VAEs with high compression ratios. In experiments, we find that SpectralNorm \citep{miyato2018spectralnormalizationgenerativeadversarial} improves the training stability more effectively than the commonly used R1 penalty \citep{mescheder2018trainingmethodsgansactually} or LeCAM regularization \citep{lecamgan}. To this end, we remove all BatchNorm layers and apply SpectralNorm to all convolutional layers in the discriminator. 
Although spectral normalization entails a tradeoff by slightly degrading quantitative reconstruction metrics compared to BatchNorm or GroupNorm \citep{wu2018groupnormalization} in the early training steps, it facilitates a more stable training process to achieve a better final reconstruction performance.

\subsection{Diffusion Transformer Model}\label{subsec:dit}

A diffusion model is employed to generate images and videos within the compact latent space of a VAE. Conceptually, diffusion models produce samples by beginning with random noise and progressively denoising it until reaching the complete data latent. Diffusion models have been a prominent class of generative models and the transformer architecture is used~\citep{vaswani2017attention,peebles2023scalable}.

Following~\citep{sora,dehghani2024patch}, we train on a mixture of images and videos at their native resolutions and durations. Both images and videos are flattened into sequences of VAE tokens with varying lengths. To balance runtime computation, shorter sequences are packed together; see \Cref{subsec:training_infra} for further discussion. In this section, we highlight the key design choices and share our observations regarding their impact.



\begin{wrapfigure}{r}{0.38\textwidth}
  \vspace{-5mm}
  \centering
  \includegraphics[width=0.9\linewidth]{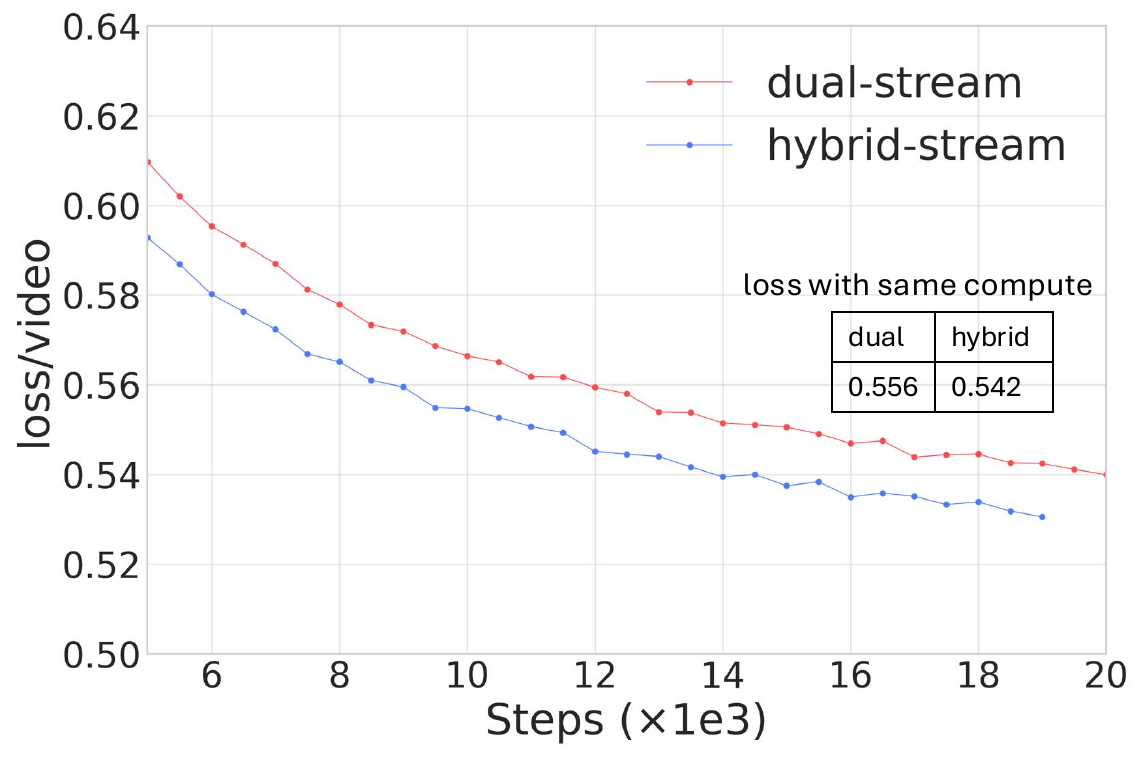}
  \caption{\label{fig:dual_stream}Loss comparison between the dual-stream and the hybrid-stream architectures. The table compares the two losses under the same training FLOPs.}
  \vspace{-5mm}
\end{wrapfigure}

\textbf{Hybrid-stream structure exhibits faster convergence.}

We employ the dual-stream DiT~\citep{esser2024scaling} as our diffusion backbone, where both video and text tokens are processed through multiple self-attentions and feedforward networks (FFNs), allowing each modality to develop its own representations. We use SwiGLU instead of GeLU as the activation function. To further improve the parameter efficiency and reduce memory cost, we use AdaSingle~\citep{chen2023pixart} for timestep modulation and share two-thirds of the FFN parameters in deeper layers. We refer to this architecture as hybrid-stream and find that it achieves faster convergence with the same number of model parameters and compute budget.

\Cref{fig:dual_stream} compares the dual-stream~\citep{esser2024scaling} and hybrid-stream architectures under the same training steps where comparison under the same compute budget is also presented.
The results indicate that the hybrid-stream architecture consistently achieves lower loss compared to the dual-stream architecture. Based on the above designs, we build the 7B hybrid-stream model with a hidden size of 3584 and a total of 32 layers.


\textbf{Full-attention enjoys training scalability.} Video generation faces challenges in long-context modeling. A 720x1280 video, just 5 seconds long at 24 fps, contains 120 frames which makes it easily exceed 100,000 tokens. This illustrates the tradeoff between attention capacity and sequence length.

\begin{wrapfigure}{r}{0.5\textwidth}
  \vspace{-5mm}
  \includegraphics[width=0.9\linewidth]{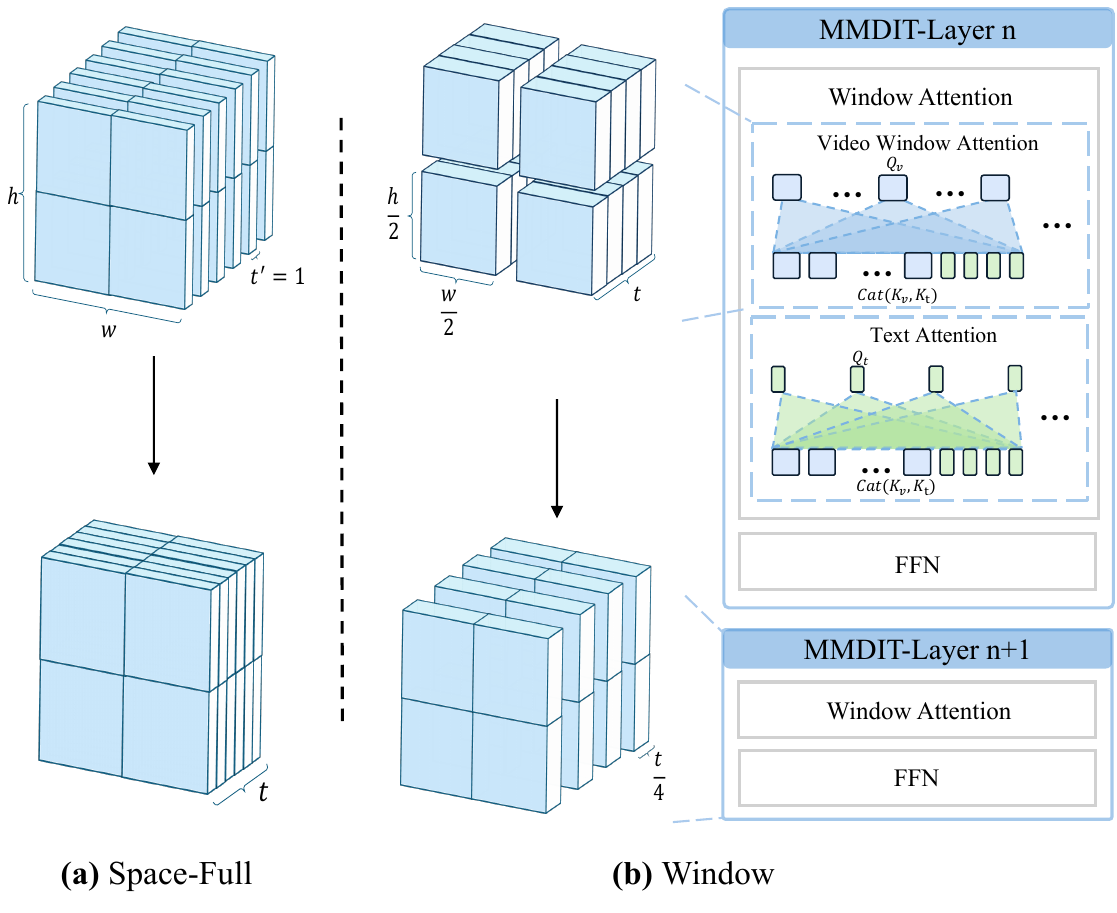}
  \caption{Illustration of the space-full and window attention architecture.}
  \label{fig:attention}
   \vspace{-1mm}
\end{wrapfigure}

We consider three types of attention: \textit{full attention}, \textit{space-full attention}, where we interleave full attention and space-only attention in every other transformer layer, and sparse \textit{window attention}, where attention is computed only over pre-defined window sets in each layer. \Cref{fig:attention} illustrates different attention architectures.

\begin{figure}[h!]
  \centering
  \begin{minipage}{0.49\linewidth}
    \centering
    \includegraphics[width=0.8\linewidth]{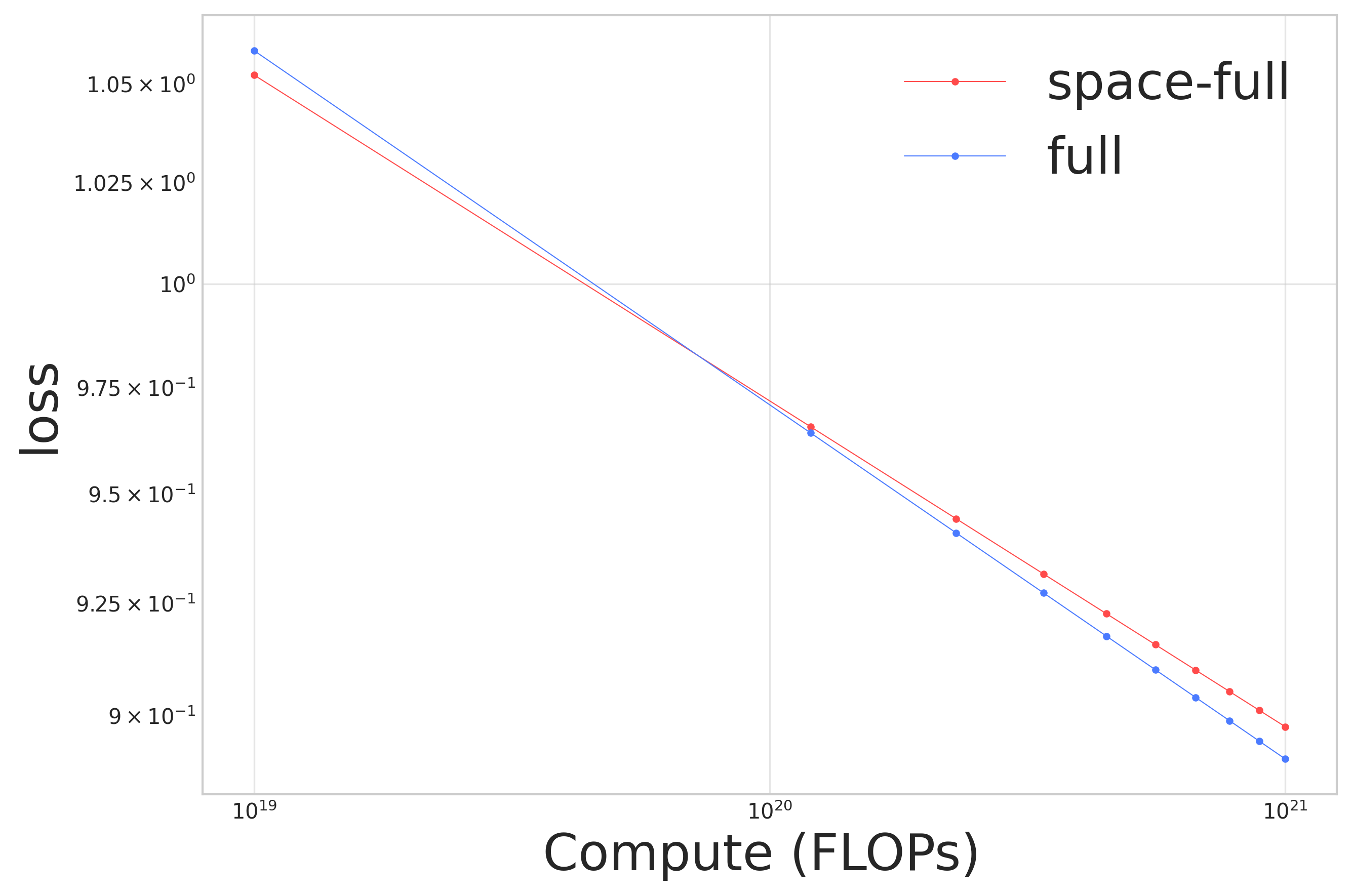}
    \caption{Loss comparison of full and space-full attention.}
    \vspace{-3mm}
    \label{fig:scaling_laws}
  \end{minipage}
  \hfill
  \begin{minipage}{0.49\linewidth}
    \centering
    \includegraphics[width=0.8\linewidth]{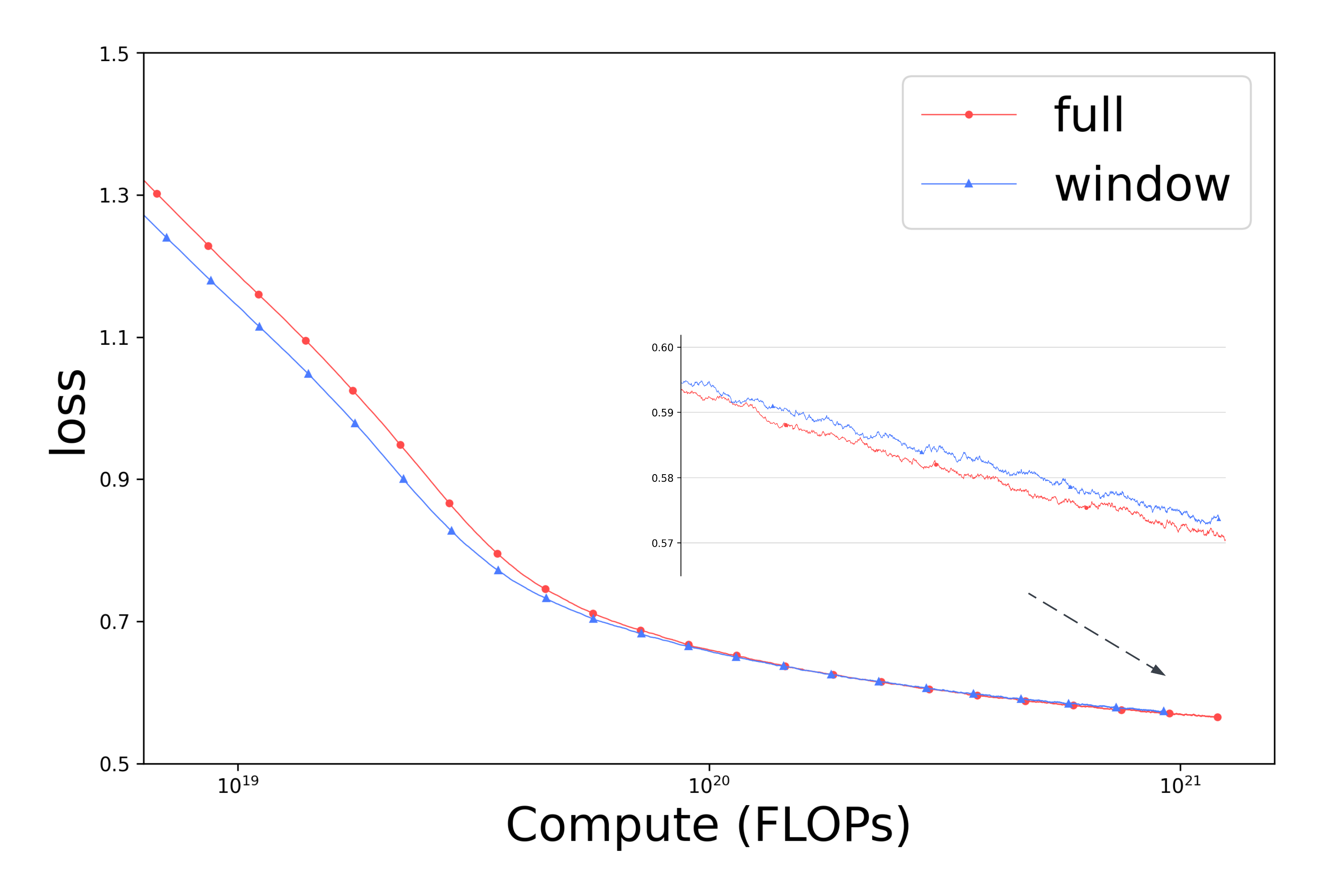}
    \caption{\label{fig:attn_loss}Loss comparison of full and window attention.}
    \vspace{-3mm}
  \end{minipage}
\end{figure}

Following~\citep{liang2024scaling, yin2024towards}, we conduct the scaling laws~\citep{liang2024scaling, yin2024towards} for \textit{full} and \textit{space-full} attention. \Cref{fig:scaling_laws} presents the power-law between the loss and compute, where a greater slope usually denotes better scalability. With a sufficient computing budget, the vanilla full attention produces a lower loss. It is worth noting that the advantage of lower loss appears to be not clearly manifested in the text-to-video task, where human raters perceive only a marginal improvement. In contrast, full attention demonstrates benefits in the image-to-video task by generating more consistent and natural motion.


As various window attention designs exist, motivated by~\citep{liu2021swin, liu2022video,gupta2024photorealistic}, we examine a simple 3D window attention mechanism by partitioning the input into $w_t \times w_h \times w_w$ windows and applying alternating attention patterns. Specifically, even-numbered layers use $1 \times 2 \times 2$, while odd-numbered layers adopt $4 \times 1 \times 1$, as illustrated in \Cref{fig:attention}. Given the constrained compute budget, such sparse window attention achieves a lower loss than \textit{full} attention. However, as the training step increases, full attention eventually surpasses window attention, as shown in \Cref{fig:attn_loss}.

These observations suggest that full attention benefits from better training scalability when given sufficient GPU resources. However, for high-resolution video training, full attention imposes a substantial computational burden, as attention computation accounts for a significant portion of the overall cost. 
To address this, a practical approach is to fine-tune the model from full to window attention after pretraining. This strategy can reduce redundancy in attention while maintaining inference efficiency, and if carefully designed, it results in negligible quality degradation.

\begin{wrapfigure}{r}{0.32\textwidth}
  \vspace{-5mm}
  \includegraphics[width=0.9\linewidth]{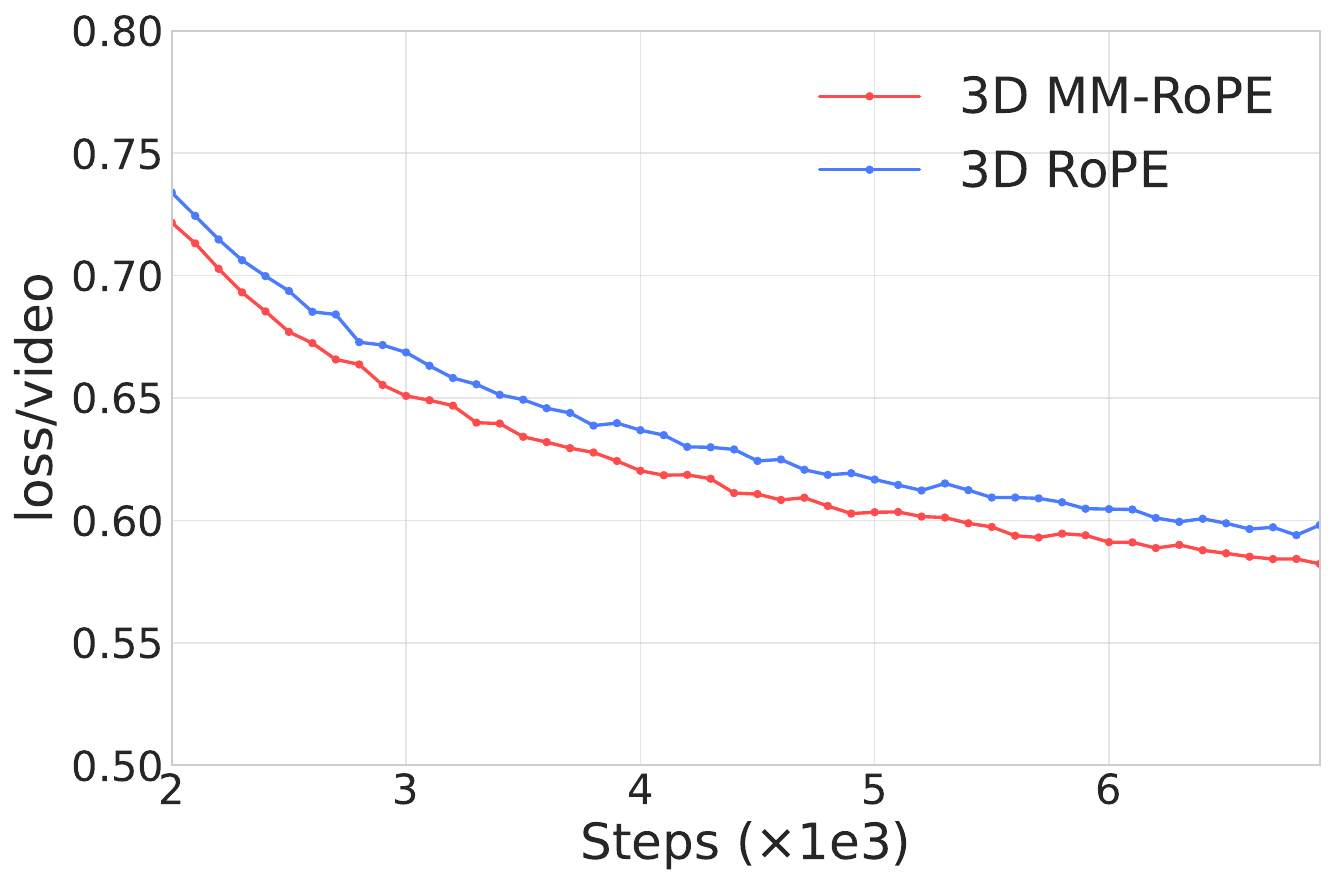}
  \caption{Loss comparison between RoPE and MM-Rope.}
  \label{fig:rope_loss}
   \vspace{-5mm}
\end{wrapfigure}

\textbf{The benefit of multimodal rotary position embedding~(MM-RoPE).} To enhance the positional information introduced by various aspect-ratio and duration, we apply 3D RoPE encoding~\cite{su2024roformer} including three components~(\ie temporal, width and height) for videos tokens, considering both the absolute and relative position dependency in attention computation. To further facilitate the effective fusion of positional information between text and videos, inspired by~\citep{wang2024qwen2},  we build 3D MM-ROPE in the concatenated sequences by adding compatible 1D positional encoding for text tokens, where three components share the same position ID. We find that this design leads to lower training loss in the dual-stream MMDiT structure.


\subsection{Training Stages}\label{subsec:training}
We use the multi-stage training strategy from low resolution to high resolution~\cite{gupta2024photorealistic,blattmann2023align}. In this section, we provide an overview of multi-task and multi-stage pre-training and post-training. Our design focuses on strategically allocating GPU resources during training to enhance overall quality.

\subsubsection{Pre-training}
We start with pre-training the model from low-resolution images only, such that the alignment between text prompts and common visual concepts can be built.

Afterward, joint image and video training is initiated, where image and video tokens are flattened and mixed at a fixed ratio within each batch. During low-resolution training, we find that incorporating a small proportion of high-resolution images allows the model to generate higher-resolution videos in a zero-shot manner, which indicates the model's ability to generalize across both modality and resolution.

The configurations for the different training stages are provided in \Cref{table:training-strategy}. There are four stages, each named after the primary target resolution area used during that phase. For example, Stage 1 primarily uses 256×256 and 512×512 images, along with 256×256 videos. Here, the resolution (\eg, 256×256) refers to the target area, not the exact dimension; the images and videos are resized while preserving their aspect ratio to match the desired area. In Stage 0, we observe that training on images alone is beneficial, in contrast to the approach of incorporating a small proportion of video as done in~\cite{gupta2024photorealistic}. Sufficient training in Stage 0 is essential for strong prompt-following capabilities.

For video training, we use multi-task training that includes text-to-video, image-to-video, and video-to-video extensions. The input features and conditioning features (\eg, the first-frame latent) are concatenated along the channel dimension, along with a binary mask indicating whether each denoised frame contains the conditioning. 

Text-to-video is the most cost-effective task for model convergence. We find that introducing a small proportion of the image-to-video task during pre-training benefits the learning of both text-to-video and image-to-video. However, excessively increasing the ratio has detrimental effects and does not improve image-to-video performance. We thus set the image-to-video ratio to 20\%. To enhance image-to-video performance, after pre-training, a dedicated image-to-video model is branched out in which the image-to-video task ratio is increased to 50-75\%. 




%
\begin{table}[h!]
  \centering
  \begin{tabular}{c|c|c|c}
    \toprule
    \textbf{Training stage} & \textbf{Image Resolution} & \textbf{Video Resolution} & \textbf{Step Percentage}\\
    \hline
    Stage 0: 256p   & [256$\times$256, 512$\times$512]  & -  & 37.5\% \\
    \hline
    Stage 1: 256p & [256$\times$256, 512$\times$512]  & [256$\times$256] & 25.0\% \\
    \hline
    Stage 2: 480p &  [640$\times$480, 1280$\times$720]  & [640$\times$480]  & 25.0\% \\
    \hline
    Stage 3: 720p & [1280$\times$720, 1920$\times$1024]  & [1280$\times$720]  & 12.5\% \\
    \bottomrule
  \end{tabular}
  \caption{Summary of the pre-training stages. Step Percentage is the proportion of total training steps allocated to each stage. The image and video resolution (\eg, 256×256) refers to the target area, not the exact dimension.}
  \label{table:training-strategy}
\end{table}

\subsubsection{Post-training}
After the pre-training phase, we apply supervised fine-tuning (SFT), followed by reinforcement learning from human feedback (RLHF), to further improve the aesthetic quality, motion consistency and structural coherence of the outputs. The post-training process is carried out independently for the text-to-video and image-to-video tasks.

\begin{figure}
    \centering
    \includegraphics[width=0.9\linewidth]{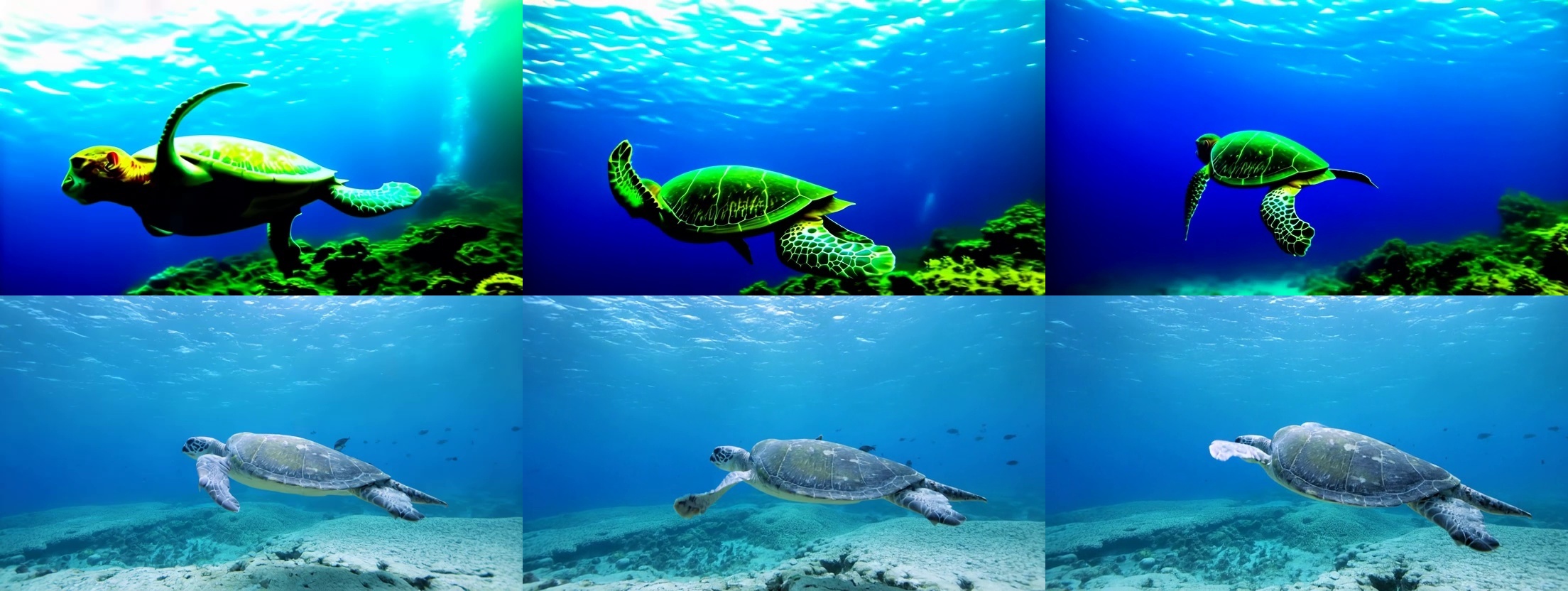}
    \caption{
    \textbf{Top}: Before SFT.
    \textbf{Bottom}: After SFT.
    Results for prompt "Turtle swimming in the ocean".
    }
    \label{fig:sft-comparison}
\end{figure}

\paragraph{SFT.} The purpose of the SFT stage is to enhance visual quality in a way that aligns more closely with human preferences, including factors like aesthetics and visual style. We curate a dataset of 700k videos with very high aesthetics and visual quality through human annotation, ensuring balanced distribution. Among these, about 50,000 videos identified as the highest quality are given greater weight during the SFT training. The SFT training is conducted on 256 GPUs using a constant learning rate, corresponding to the pre-training stage's final learning rate. As shown in \Cref{fig:sft-comparison}, after the SFT stage, the aesthetics and color quality of the generated videos are significantly improved. 

\begin{figure}
    \centering
    \includegraphics[width=0.9\linewidth]{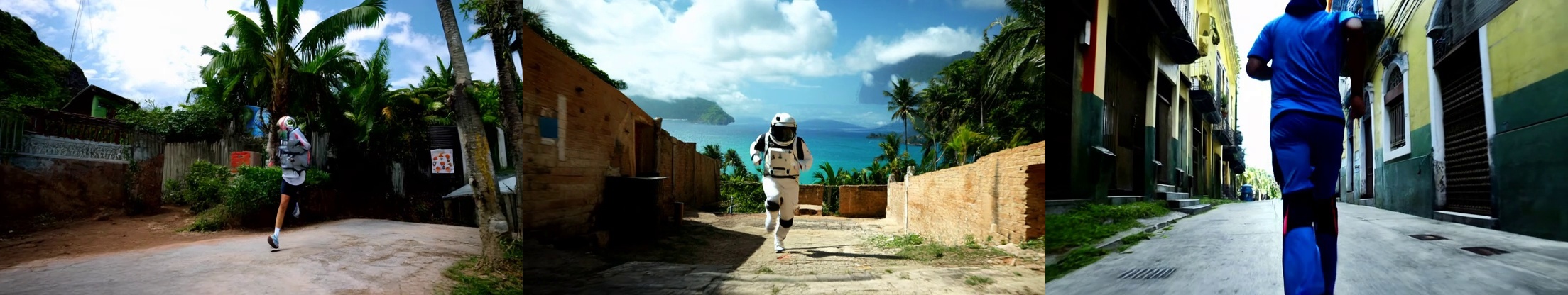}
    \caption{
    \textbf{Left}: Before SFT.
    \textbf{Middle}: Good SFT.
    \textbf{Right}: Overfit SFT.
    Results for prompt: "An astronaut running through an alley in Rio de Janeiro, 4k, high resolution".
    }
\vspace{-3mm}    
    \label{fig:sft-three-stages}
\end{figure}

However, prolonged SFT training can lead to rapid overfitting, resulting in degraded prompt-following ability and reduced motion quality. As illustrated in \Cref{fig:sft-three-stages}, the model’s ability to follow prompts deteriorates  during the SFT stage.


\begin{figure}
    \centering
    \includegraphics[width=0.9\linewidth]{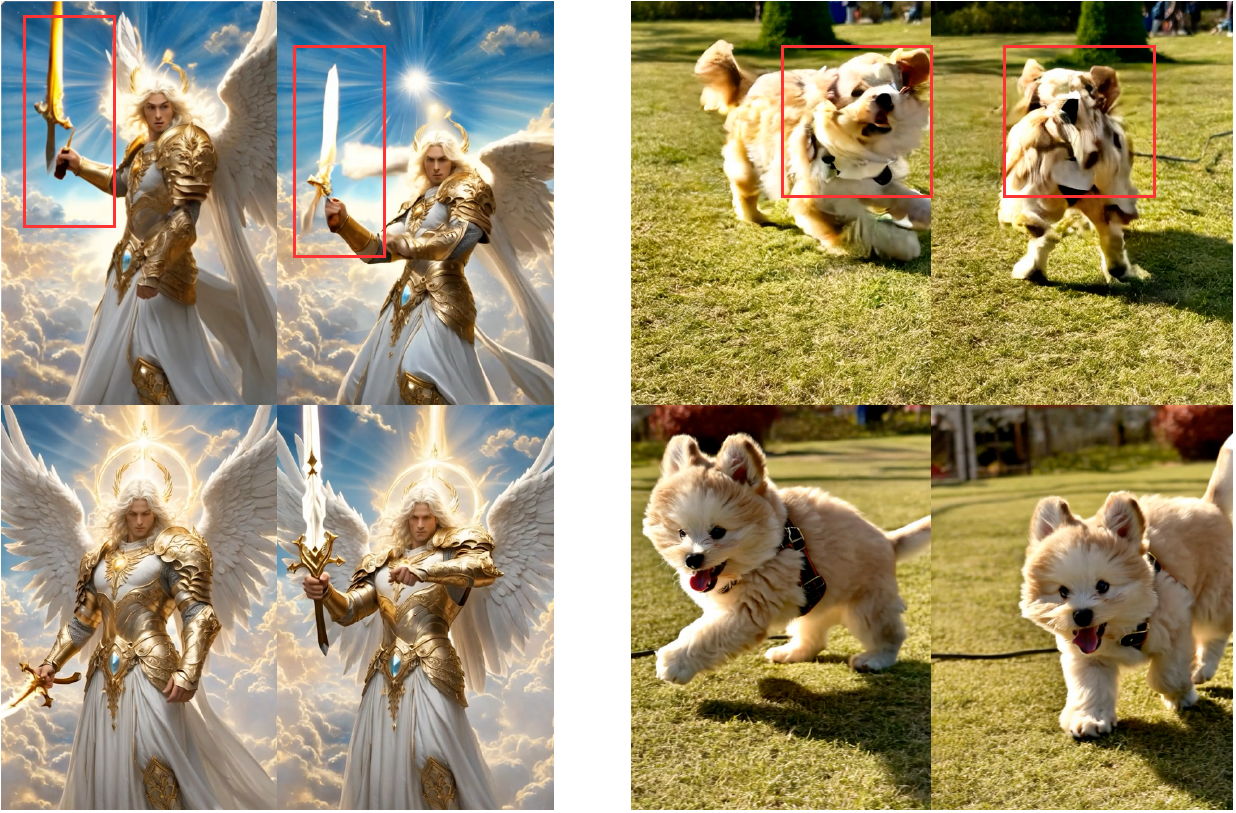}
    \caption{Two image-to-video examples before (top row) and after (bottom row) DPO. DPO significantly improves the structure and motion quality.}
    \label{fig:dpo_comparison}
\end{figure}

\paragraph{RLHF.}
Though SFT is effective in improving aesthetics, degradation in motion and structure is frequently observed after SFT. We find that Direct Preference Optimization (DPO)~\cite{rafailov2023direct} is highly effective in addressing these issues. Inspired by the simplified loss function in Diffusion-DPO~\cite{wallace2024diffusion}, we develop a DPO approach for video generation, which incorporates SFT loss on positive samples during training.
Empirically, we set an extremely small learning rate as $1e^{-7}$ ($50\sim 100\times$ smaller than SFT) and a large $\beta=100$ (see Eqn. 14 in \cite{wallace2024diffusion}). We collect video-text pairs from our pretraining and SFT datasets which are balanced by textual and visual clustering. For each video-text pair, we generate 4 videos and ask annotators to select the best and worst videos among them. As shown in \Cref{fig:dpo_comparison}, DPO is highly effective in improving the structure and motion quality.

\noindent \textbf{Switching reference model is not necessary.}
In LLaMA~\cite{grattafiori2024llama}, multi-round DPO is performed by switching the reference model on the same set of preference data. However, in our case, this approach results in only marginal improvements in the target dimensions (\eg, structure collapse) and noticeable downgrades in other dimensions (\eg, prompt following, color correctness). Consequently, we employ a consistent reference model (\eg, the SFT model) and do not use the multi-round DPO.

\noindent \textbf{Specialization for the image-to-video task.} In the image-to-video task, the first frame of the generated video should remain consistent given the same prompt (conditioning image and text prompt). This renders the loss function 
less suitable, as it maximizes the distance between the latent of the first frame of the positive and negative videos, which should be identical. Specifically, we observe that using the conventional DPO leads to oversaturation of the first frame. To address this issue, we separate the diffusion loss computation for the first-frame latent and apply the DPO loss to the latent of the subsequent frames.

\begin{figure}[h!]
  \centering
  \includegraphics[width=\linewidth]{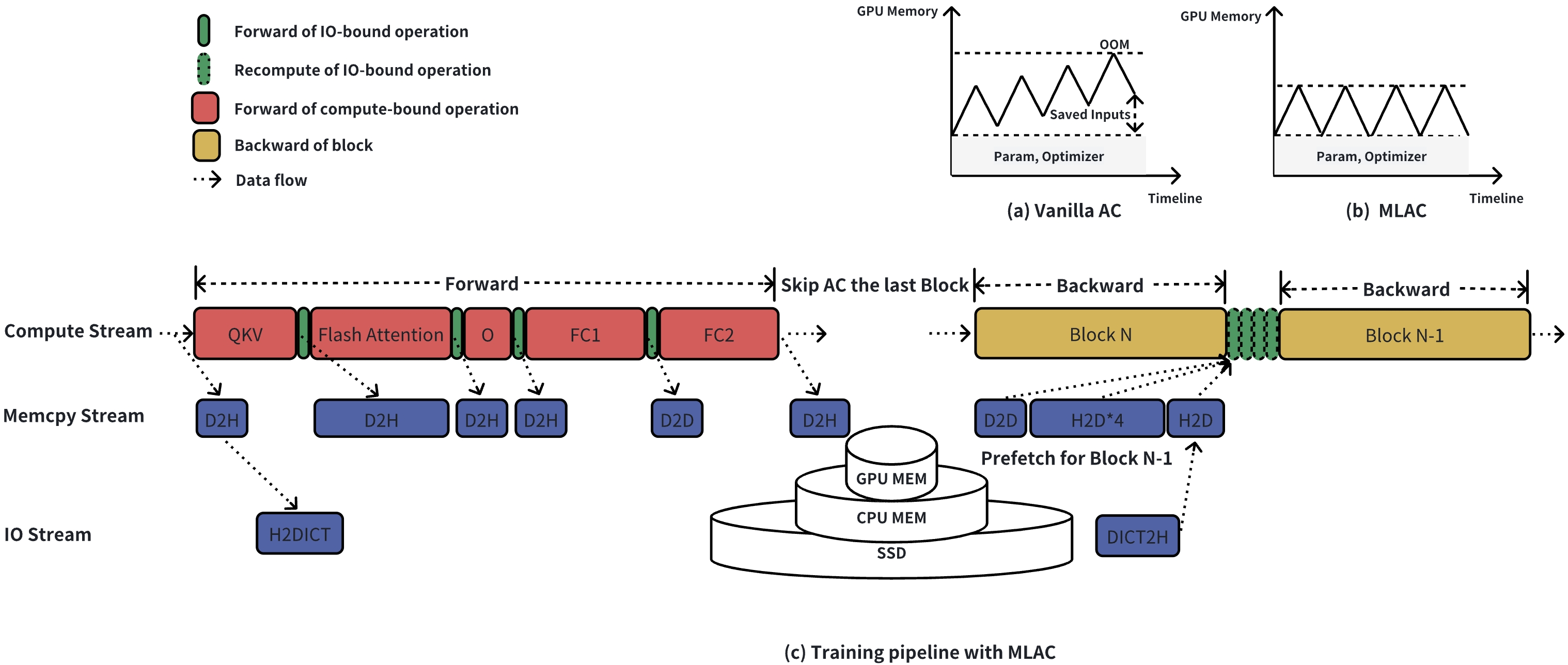}
  \caption{Multi-level activation checkpointing(MLAC). (a) Vanilla AC saves inputs on device could still encounter GPU OOM. (b) MLAC further supports offloading module inputs to achieve zero-activation AC, (c) and minimize recomputation overheads by saving compute-bound activations to multi-level storage space.}
  \label{fig:mlao}
\end{figure}

\subsection{Infrastructure}
\label{subsec:training_infra}
This subsection outlines the key strategies for efficient model training. First, we employ a parallelism strategy to train a 7B model on long-context videos. Second, we introduce a Runtime Balance strategy to mitigate load imbalance during joint image-video training. Additionally, we design Multi-Level Activation Checkpointing (MLAC) to reduce GPU memory usage and recomputation overhead. Finally, we optimize GPU utilization by implementing fused CUDA kernels to streamline fragmented I/O operations. As a result, \modelname{} achieves a Model FLOPs Utilization (MFU) of 38\% in distributed training at large-scale.

\begin{figure}[h!]
  \centering
  \includegraphics[width=.8\linewidth]{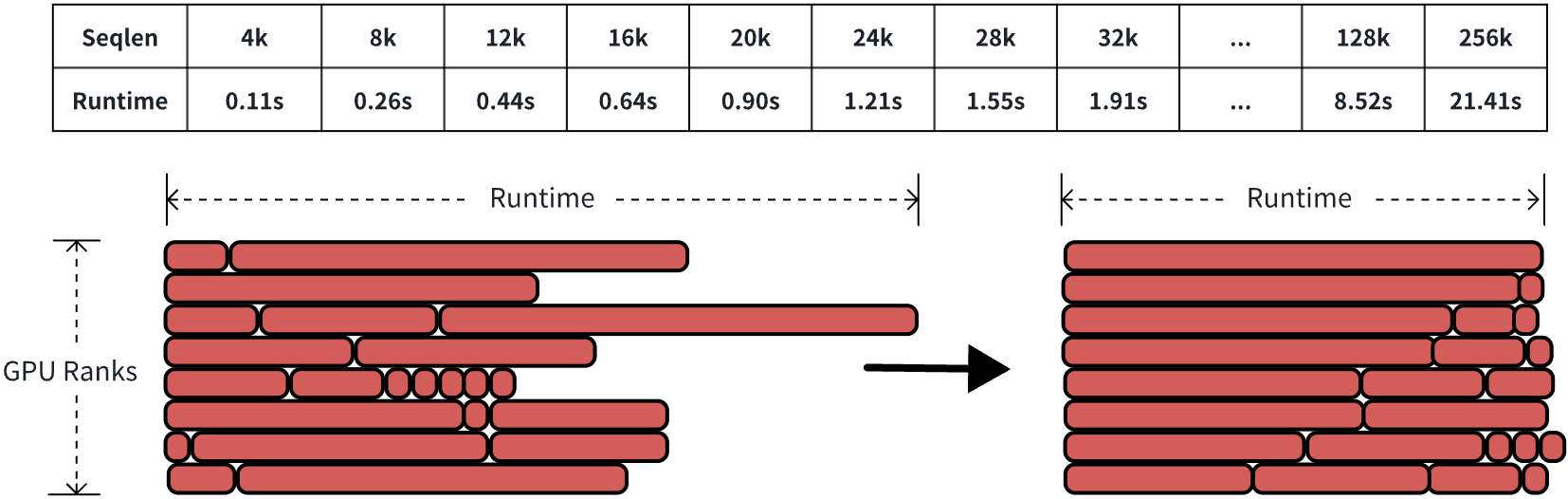}
  \caption{Balance samples within one batch across GPUs by runtime metric.
        \textbf{Top}: seqlen-to-runtime lookup table.
        \textbf{Bottom left}: One batch samples across GPUs before balance.
        \textbf{Bottom left}: One batch samples across GPUs after balance.}
  \label{fig:runtime_balance}
\end{figure}

\paragraph{Parallelism Strategy.}
We use 3D parallelism including data-parallelism, context-parallelism, and model sharding to support efficiently distributed training with long-context samples. FSDP~\citep{zhao2023pytorch} shards the model parameters, optimizer states, and gradients across GPUs, and enables overlap between computation and communication to reduce communication overhead and improve distributed training efficiency. We adopt Ulysess~\citep{jacobs2023deepspeed} as our context-parallelism strategy, which iteratively shards samples across the sequence dimension and the head dimension in token-dependent/independent layers through all-to-all communication.

\paragraph{Runtime Balance.}
Joint training of images and videos causes significant load imbalance across GPUs, leading to inter-device synchronization overhead and reduced training throughput. Existing methods~\citep{ma2025step} based on seqlen(sequence length) and FLOPs fail to achieve optimal balance due to the nonlinear relationship between training runtime and these metrics, influenced by varying operator efficiency. To address this, we propose Runtime Balance (\Cref{fig:runtime_balance}), which constructs an offline lookup table mapping seqlen to actual runtime. During training, runtime estimates are obtained via table queries, ensuring optimal workload distribution. Runtime balancing is restricted to ranks within the same batch to preserve data consistency. To minimize overhead, balancing for the next batch is performed asynchronously in a sub-process, preventing delays in the main training process.

\paragraph{Multi-Levels Activation Checkpointing.}
Activation checkpointing~\citep{chen2016training} introduces significant recomputation overhead during backpropagation and may still encounter GPU OOM issues in long-context scenarios, as it inevitably requires caching the input tensor of the wrapped module. We introduce Multi-Level Activation Checkpointing (MLAC), as shown in \Cref{fig:mlao}, a method for selectively saving any intermediate activation during the forward pass on multiple levels of storage, such as GPU, CPU, and disk memory. MLAC minimizes recomputation overhead during backpropagation by prioritizing the caching of output tensors from compute-bound operations. Furthermore, it supports offloading input tensors of the gradient checkpointing module on the CPU and disk to attain zero
activation occupancy in GPU memory, which allows for training larger models with longer context.
MLAC incorporates an efficient asynchronous caching and prefetching mechanism that optimizes the overlap between memory transfers and forward/backward computations. Compared to the vanilla AC, MLAC makes full use of available hardware resources and significantly enhances training efficiency.

\paragraph{Fused Kernel.}
IO-bound operations such as normalization and RoPE (rotary position encoding) frequently access memory, preventing the Tensor/CUDA core from being fully utilized. 
We introduce kernel fusion techniques that leverage registers and shared memory to store intermediate results from consecutive memory-access-intensive operators and fuse them into a single CUDA kernel. These fused kernels reduce the global memory accesses to one-tenth of the baseline, significantly improving the kernel's arithmetic intensity. Specifically, we fuse QK-Norm, RoPE, and all the attention preprocessing operations and implement the corresponding forward and backward fused kernels. Similar optimization strategies are applied to the rest of the memory-access-intensive operators in the model to improve the training and inference efficiency. 


\subsection{Optimizations}\label{subsec:optim}

\paragraph{Inference optimizations.} To accelerate model inference, we use diffusion distillation techniques to reduce the number of function evaluations (NFE) required for generation. Our acceleration process consists of three stages. First, we adopt the trajectory segmented consistency distillation (TSCD) method proposed in HyperSD~\cite{renhyper}, enabling the model to perform satisfactorily at approximately 24 NFE. Then, we design a classifier-free guidance (CFG) \cite{ho2022classifier} embedding module and perform CFG distillation to eliminate the inefficiencies caused by the two-NFE-per-step inference process in CFG while maintaining guidance scale parameterization. Our embedding module supports the input of both the CFG scale and negative prompts. Finally, to mitigate the blurriness introduced by few-step inference, we conduct adversarial training with the model fixed at 8 NFE. The details of the adversarial training design, as well as further improvements in 1–2 NFE performance, can be found in the Seaweed-APT paper~\cite{lin2025diffusion}. Based on this three-stage distillation scheme, our distilled 8-NFE model achieves comparable performance to the original model in text alignment and motion quality and even demonstrates superior results in visual fidelity. For example, in a representative evaluation, the 8 NFE model maintains a competitive win rate (56\%) compared to the original model (58\%).

\paragraph{VAE optimizations.}
VAE's causal chunking significantly reduces memory consumption. This supports encoding and decoding of videos at up to 1280$\times$720 resolution of any length on a single GPU with 40+GB of memory. For tasks requiring higher resolutions, we split the feature map into smaller sections for convolution and normalization layers, thereby reducing peak GPU memory usage.
For further speedup, our VAE employs a multi-GPU pipeline. We split the video along the temporal dimension and distribute the segments across multiple GPUs, achieving sequence parallelization. Specifically, each GPU processes a consecutive chunk, and every causal convolution layer sends the sliced padding cache to the next GPU.

\paragraph{Rephraser.}
We find that using captions from professional training videos as input during DiT inference improves visual aesthetics and motion stability.
Therefore, we train a model to rephrase the input prompts from the user to align with the style of captions from high-quality videos in our training set.
To do so, we first curate a parallel corpus by pairing the simulated with the input prompts with detailed video captions. 
Then, a 7B Large-Language Model (LLM) is fine-tuned to convert the input prompts into detailed captions. 
To mitigate semantic drift, the model generates 8 variants per prompt following Supervised Fine-Tuning (SFT). Semantically accurate variants are selected as positive samples, while semantically inconsistent ones serve as negative samples. Direct Preference Optimization (DPO) is then applied to reinforce outputs that balance accuracy and quality.

Our final rephrase model notably enhances video generation, especially in terms of visual aesthetics and style. However, it can compromise prompt following, particularly for longer input prompts (over twelve words), where preserving the exact semantic meaning during rephrasing becomes challenging.



%% file: sections/4_evaluation.tex
\section{Evaluation}

In this section, we evaluate \modelname{}'s generation capabilities as a video generation foundation model. Specifically, in \cref{sec:exp_dit}, we empirically compare \modelname{}'s performance in text-to-video and image-to-video generation with the contemporary models. \cref{sec:vae} presents an analysis of our VAE against state-of-the-art VAE models in terms of reconstruction quality. Finally, \cref{sec:exp_inf} discusses the inference process of the generation model.

\subsection{Quantitative Analysis of Video Generation}
\label{sec:exp_dit}

To evaluate generic generation quality, we examine two tasks: image-to-video and text-to-video. Although text-to-video generation is a primary focus in current literature, we note that image-to-video is far more popular among users.

To assess the overall ranking among competing models, we use MagicArena (\url{https://aigcarena.com/}), which employs an Elo rating system~\citep{elo1978rating}. Note that we intentionally choose human evaluation, as the community has yet to develop automatic metrics capable of assessing video generation quality at the level of human raters. In the Elo framework, two videos generated by different models from the same prompt are randomly selected and displayed side by side. Human raters evaluate the videos based on their overall fidelity.
The raters can either choose the superior clip or indicate a tie. A model that consistently outperforms its peers accumulates a higher Elo score and a higher win ratio. More than 500 raters participated in the evaluation, with each model involved in at least 7,000 pairwise comparison trials. The baseline videos were generated using the models' latest official APIs as of early March 2025. For more details, please refer to MagicArena.

\begin{table}[h]
\centering
\begin{tabular}{@{}lccc@{}}
\toprule
\textbf{Name} & \textbf{ELO} & \textbf{Win Ratio} & \textbf{Model Size}\\ 
\midrule
Kling 1.6 HD                   & 1,065 & 61\% & -\\
\underline{Seaweed-7B}          & 1,047 & 58\% & 7B\\ 
Wan 2.1         & 1,015 & 53\% & 14B\\
Luma Ray 2        & 1,003 & 51\% & -\\
Runway Gen-3 Alpha        & 1,000 & 53\% & -\\
Veo 2.0                   & 992 & 50\% & -\\
HunyuanVideo                  & 944 & 43\% & 13B \\ 
Sora                   & 903 & 36\% & -\\ 
\bottomrule
\end{tabular}
\caption{Elo comparison for the image-to-video task.}
\label{tab:elo_i2v}
\end{table}

\Cref{tab:elo_i2v} presents the results for the image-to-video task, where the \modelname{} model ranks second, outperforming several contemporary strong models such as Sora~\cite{sora}, and Veo 2.0~\cite{veo2}.
This result is particularly significant as \modelname{}, a 7B model trained with computational resources equivalent to training on 1,000 H100 GPUs over 27.7 days, surpasses the performance of larger models, many of which were trained with substantially greater GPU resources, such as Wan 2.1's 14B~\cite{wan2.1} and HunyuanVideo's 13B~\cite{kong2024hunyuanvideo}.

Since Elo scores only reflect overall rankings, we perform a fine-grained comparison across key evaluation metrics. In each assessment, human raters compare two models side by side based on specific criteria. For image-to-video, the criteria are: \textit{Visual Quality}: realism, level of detail, and overall aesthetic appeal. \textit{Motion Quality}: free of collapse or artifacts, degree of motion, naturalness (adherence to physical laws). \textit{Prompt Following}: how well the video adheres to both the visual and motion elements described in the text prompt. \textit{Reference-Image Consistency}: how accurately the generated video maintains consistency with the image prompt. 

\begin{figure}[h!]
  \centering
  \includegraphics[width=0.9\linewidth]{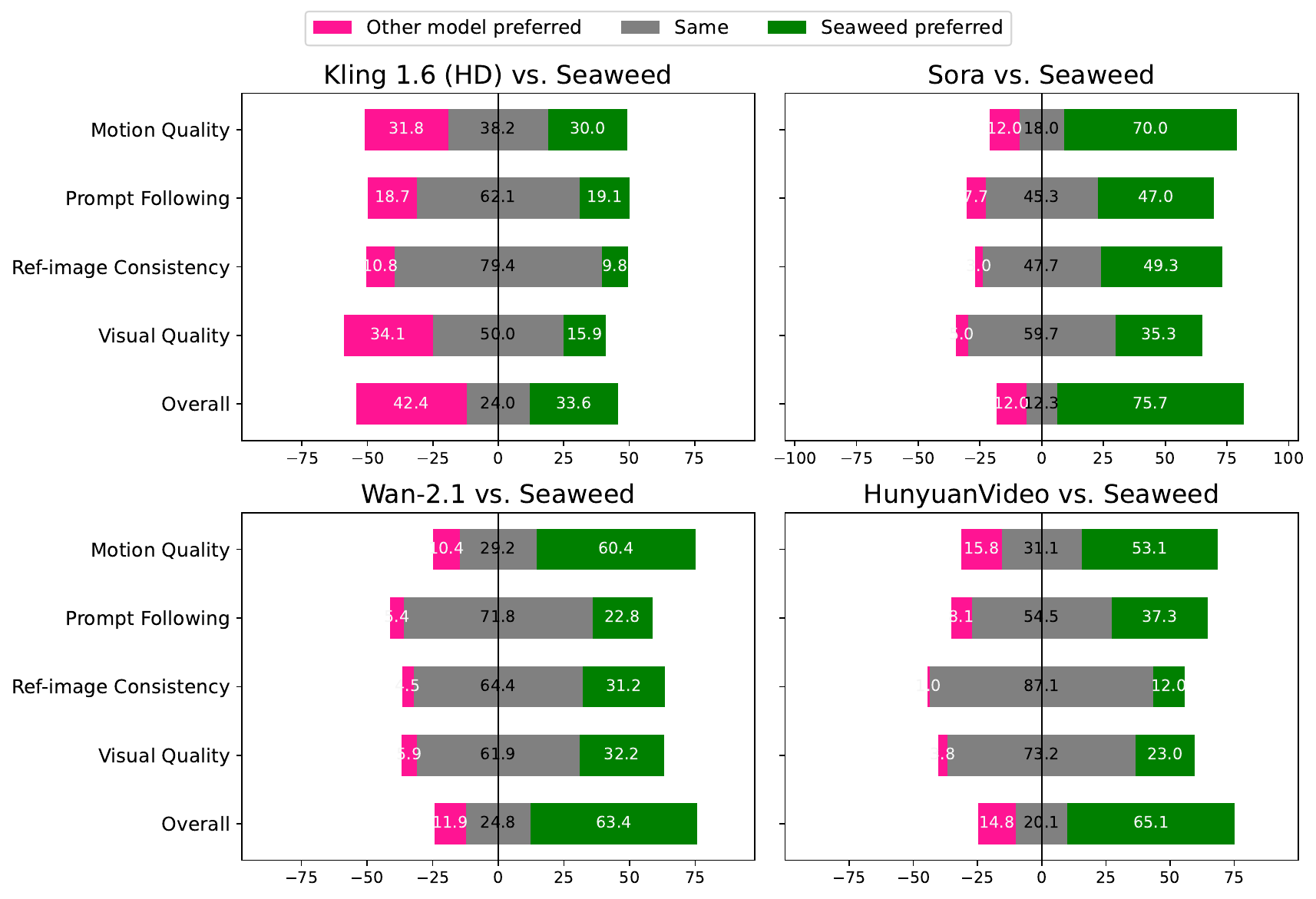}
  \caption{Comparison of \modelname{} with the top ranking models: Kling 1.6, Sora, HunyuanVideo, and Wan-2.1. The task is image-to-video.}
  \label{fig:gsb}
\end{figure}

\Cref{fig:gsb} presents a fine-grained comparison with leading baseline models. As shown, our model outperforms Sora, Wan-2.1, and HunyuanVideo by a large margin across metrics. Compared with Kling 1.6 (HD), our model is competitive in prompt following and motion quality, while it falls behind in visual quality, resulting in a lower overall ranking and the Elo ranking in \Cref{tab:elo_i2v}. This deficiency in visual quality is expected, as our results used in this test are in 480p or 720p, whereas Kling's outputs are in 1080p giving it a clear advantage in visual fidelity.

\begin{figure}[h!]
  \centering
  \includegraphics[width=0.9\linewidth]{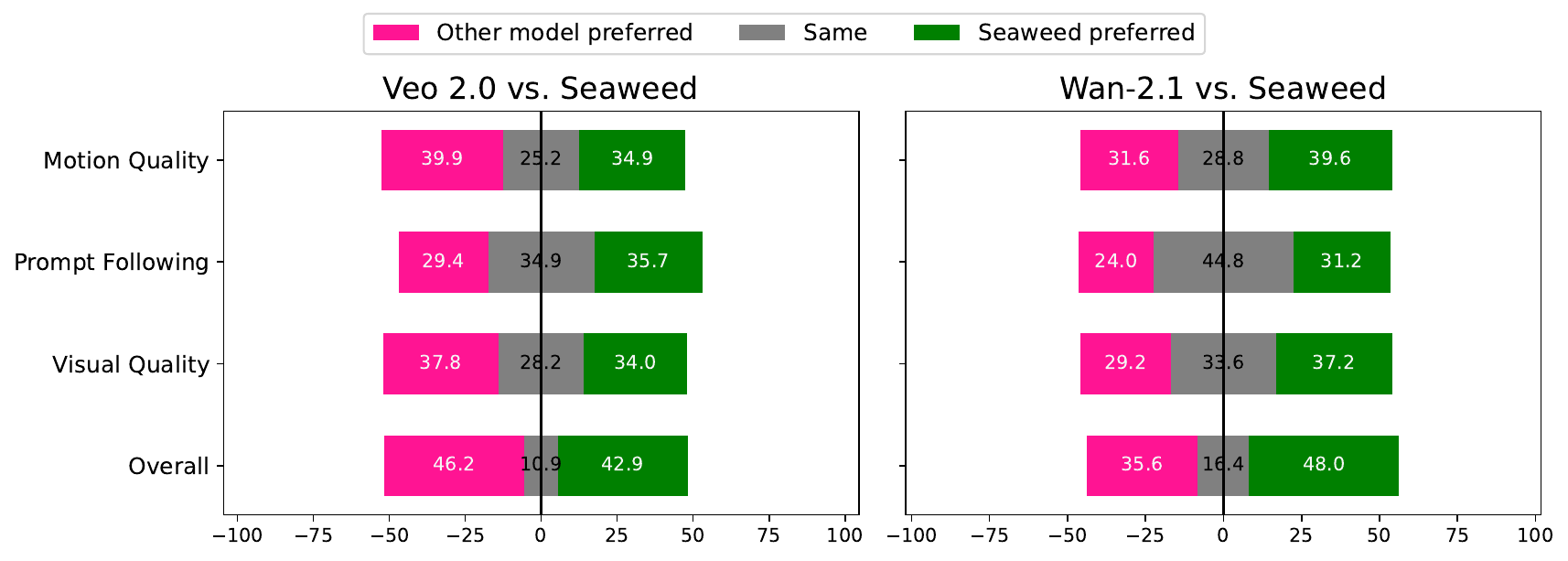}
  \caption{Comparison of the \modelname{} Model top ranking models: Veo 2.0 and Wan-2.1. The task is text-to-video.}
  \label{fig:gsb_t2v}
\end{figure}

In the text-to-video task, \modelname{} ranks among the top 2 models in the Elo comparison. It follows the top-ranked model, Veo 2, and outperforms Wan 2.1-14B and Kling 1.6 (HD). \Cref{fig:gsb_t2v} presents a detailed comparison with the two leading models, Veo 2.0 and Wan 2.1. These results demonstrate that the Seaweed model, trained with 665,000 H100 GPU hours, achieves competitive performance compared to larger models trained with significantly greater computational resources.


\subsection{Inference Time}
\label{sec:exp_inf}
In \Cref{tab:computational_efficiency}, we evaluate the inference efficiency of our model against a representative baseline with available inference metrics -- Wan-2.1~\cite{wan2.1}. The reported time represents the end-to-end generation process on a single H100 GPU, including both the text encoder and the VAE decoder modules. We evaluate Wan-2.1 using its default configuration, which employs 50 inference steps with classifier-free guidance, resulting in a total of 100 neural function evaluations (NFEs). In contrast, our model is distilled to require only 12 NFEs. Our results demonstrate that our model not only achieves superior generation quality but also operates 62 times faster.

\begin{table}[h]
    \centering
    \renewcommand{\arraystretch}{1.3}
    \begin{tabular}{l r r r r}
        \toprule
        \textbf{Model} & \textbf{Parameters} & \textbf{I2V Win Ratio} & \textbf{NFEs} & \textbf{Time (s)} \\
        \midrule
        Wan 2.1 & 14B & 53\% & 100 & 1837.9 \\ 
        Seaweed & 7B & \textbf{58\%} & \textbf{12} & \textbf{29.6} \\
        \bottomrule
    \end{tabular}
    \caption{Computational efficiency comparison between ours and Wan 2.1 measured on a single H100 GPU.}
    \label{tab:computational_efficiency}
\end{table}

\subsection{VAE Results}
\label{sec:vae}

We compare the reconstruction results of our 3D VAE model with those of state-of-the-art VAE models. We evaluate primarily using reconstruction FVD (rFVD)~\citep{unterthiner2019accurategenerativemodelsvideo}, learned perceptual image patch similarity (LPIPS)~\cite{zhang2018unreasonableeffectivenessdeepfeatures}. In addition, peak signal-to-noise ratio (PSNR) and structural similarity index (SSIM)~\cite{DBLP:journals/tip/WangBSS04} are reported for reference. For the validation set, we use the common video benchmark UCF-101 and the video compression benchmark MCL-JCV~\cite{wang2016mcl}. For UCF-101, following~\cite{yu2023magvit,yu2023language}, we select the center clip of 17 frames, resize the video to a short side of 256, and then center-crop to $256 \times 256$ for evaluation. For MCL-JCV, we choose 30 long videos (117-149 frames) at their original resolution ($720 \times 1080$) for reconstruction.

\begin{table}[h!]
\centering
\resizebox{.95\textwidth}{!}{%
\begin{tabular}{@{}lccccccccc@{}}
\toprule 
          & \multicolumn{4}{c}{} & \multicolumn{4}{c}{UCF-101} & MCL-JCV \\
        \cmidrule(lr){6-9} \cmidrule(lr){10-10}
        
          & Params (M) & $(d_t, d_h, d_w)$ & $C$ & $r$ & rFVD $\downarrow$ & LPIPS $\downarrow$ & PSNR $\uparrow$ & SSIM $\uparrow$ & LPIPS $\downarrow$ \\
\midrule
Open-Sora v1.2 \citep{opensora} & 393.3      & (4, 8, 8)     & 4  & 1:192        & 47.04 & 0.1661 & 27.70 & 0.8893 & 0.2687 \\
LTX-Video \citep{HaCohen2024LTXVideo} & 935.0      & (8, 32, 32)     & 128 & 1:192      & 45.08 & 0.1257 & 29.30 &  0.8591 & 0.2486   \\
\hline 
Cosmos ($48\times$) \citep{agarwal2025cosmos}   & 90.2       & (4, 8, 8)     & 16 & 1:48        & 13.02 & 0.0847 & 32.34 & 0.9484 & 0.1851 \\
SVD \citep{blattmann2023stablevideodiffusionscaling}      & 97.7       & (1, 8, 8)     & 4  & 1:48        & 11.10 & 0.0751 & 30.81 & 0.9356 & 0.1137 \\
Wan-VAE \citep{wan2.1}  & 126.9      & (4, 8, 8)     & 16 & 1:48      & 2.08 & 0.0463 & 34.00 & 0.9603 & 0.1034   \\
CV-VAE (SD3) \citep{zhao2024cvvaecompatiblevideovae}    & 181.9      & (4, 8, 8)     & 16 & 1:48        & 6.50  & 0.0589 & 33.21 & 0.9612 & 0.1437 \\
CogVideoX \citep{yang2024cogvideoxtexttovideodiffusionmodels} & 215.6      & (4, 8, 8)     & 16 & 1:48        & 6.06  & 0.0623 & 34.30 & 0.9650 & 0.1378 \\
HunyuanVideo \citep{kong2024hunyuanvideo}  & 246.5      & (4, 8, 8)     & 16 & 1:48        & \textbf{1.79}  & 0.0456 & 35.15 & 0.9713 & 0.1102 \\
our Seaweed VAE ($48\times$) & 250.6      & (4, 8, 8)     & 16 & 1:48        & 1.85  & 0.0517 & 33.83 & 0.9643 & 0.1477  \\
WFVAE \citep{li2024wf}     & 317.1      & (4, 8, 8)     & 16 & 1:48        & 3.15  & 0.0643 & 34.13 & 0.9687 & 0.1572 \\
\hline 
Our Seaweed VAE ($64\times$) & 552.8      & (4, 16, 16)    & 48 & 1:64        & 2.43  & \textbf{0.0391} & \textbf{35.23} & \textbf{0.9717} & \textbf{0.0945} \\
\bottomrule
\end{tabular}
}
\caption{Reconstruction results for VAE on \textbf{UCF-101} 
 (17$\times$256$\times$256) and \textbf{MCL-JCV} ([117-149] $\times$720$\times$1080) }
 \label{fig:vae-reconstruction}
\end{table}

As shown in \Cref{fig:vae-reconstruction}, our VAE achieves state-of-the-art reconstruction performance across all metrics, including rFVD, LPIPS, PSNR, and SSIM. Notably, for real-world uncompressed videos with high resolution and long duration in MCL-JCV, we achieve the lowest LPIPS, even with a higher compression ratio than the previous state-of-the-art VAE model.

\begin{figure}[h]
  \centering
  \begin{subfigure}[b]{0.49\textwidth}
    \centering
    \includegraphics[width=\textwidth]{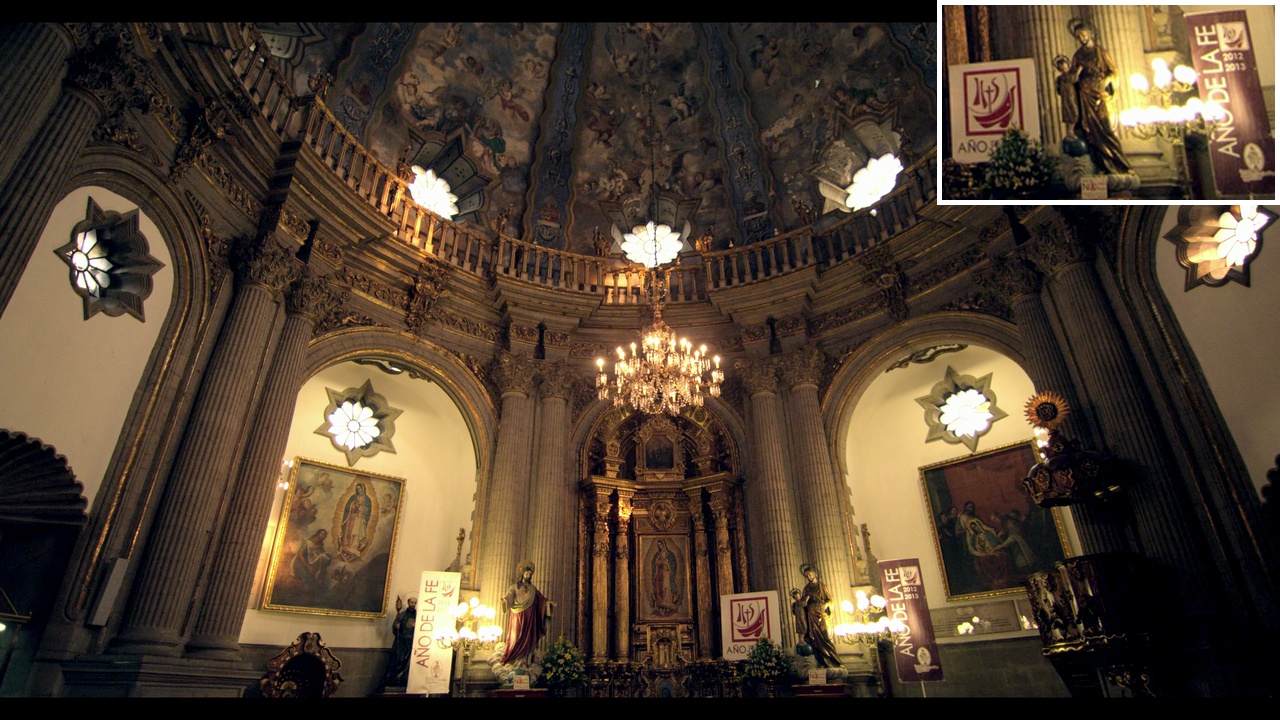}
    \caption{Original Video}
    \label{fig:vae_vis_original_1}
  \end{subfigure}
    \hfill
  \begin{subfigure}[b]{0.49\textwidth}
    \centering
    \includegraphics[width=\textwidth]{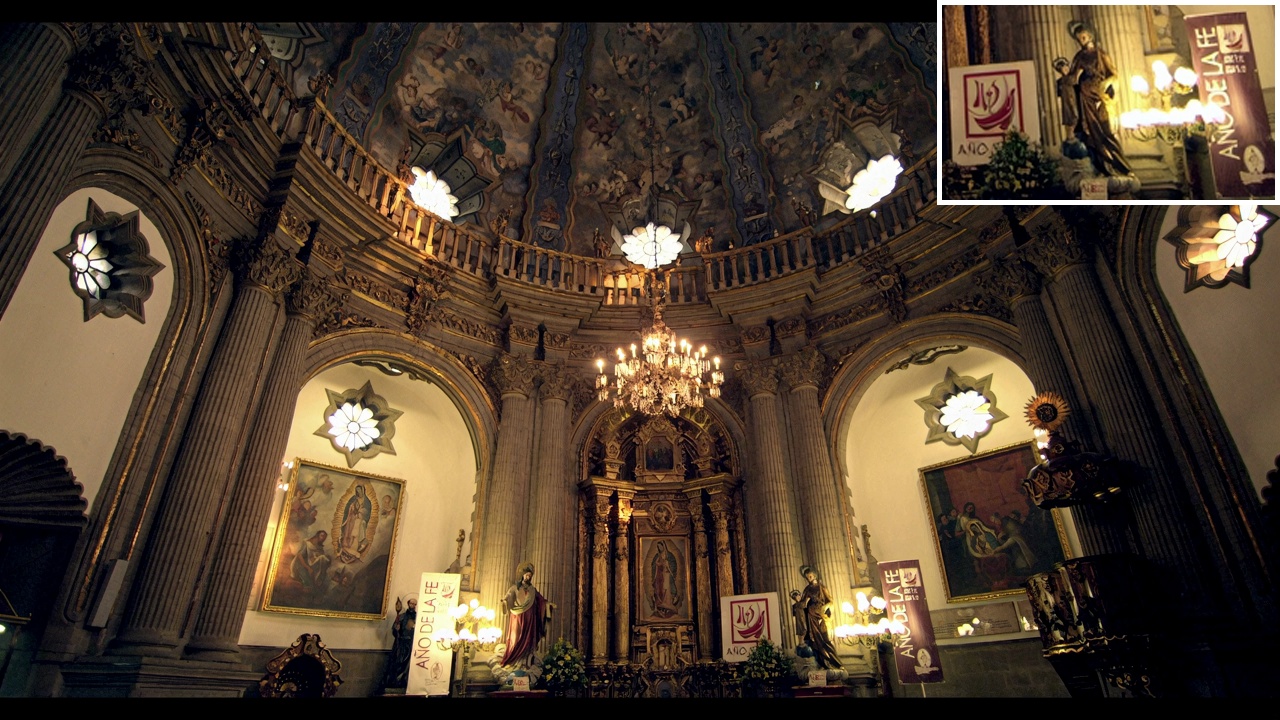}
    \caption{$48\times$ Seaweed VAE}
    \label{fig:vae_vis_seaweed1_1}
  \end{subfigure}
  \hfill
  \begin{subfigure}[b]{0.49\textwidth}
    \centering
    \includegraphics[width=\textwidth]{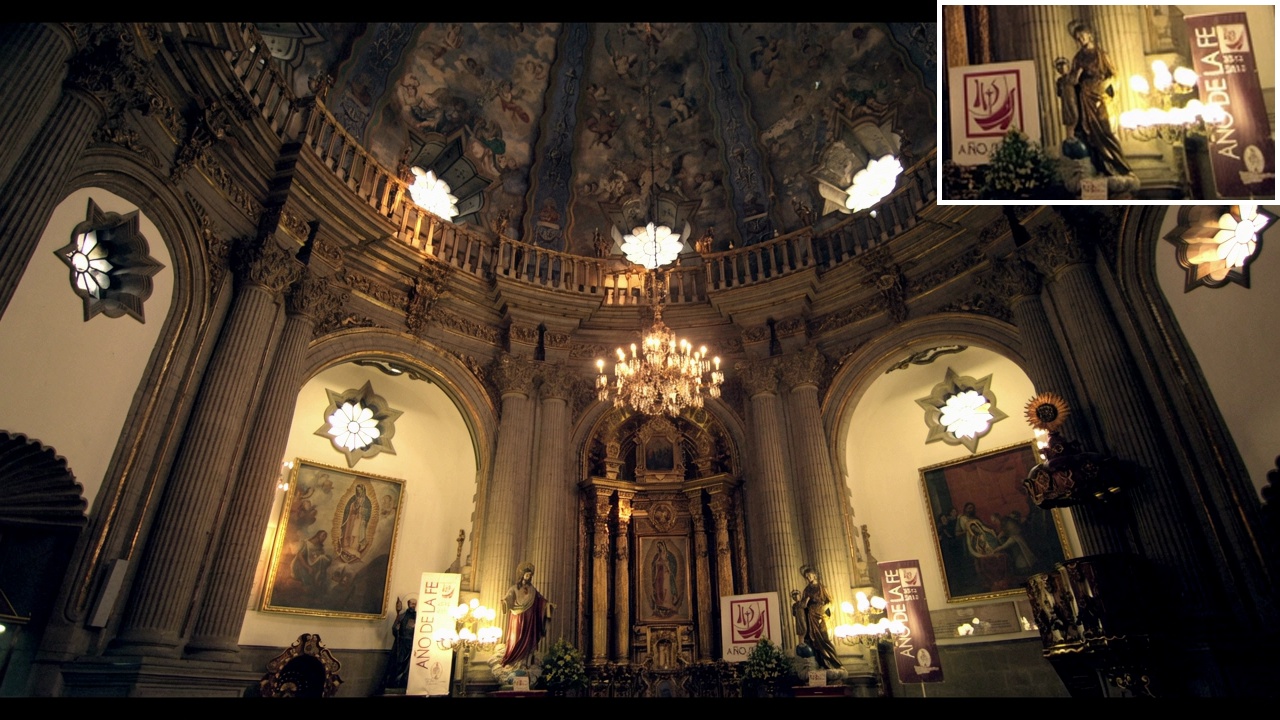}
    \caption{$64\times$ Seaweed VAE}
    \label{fig:vae_vis_seaweed2_1}
  \end{subfigure}
  \hfill
  \begin{subfigure}[b]{0.49\textwidth}
    \centering
    \includegraphics[width=\textwidth]{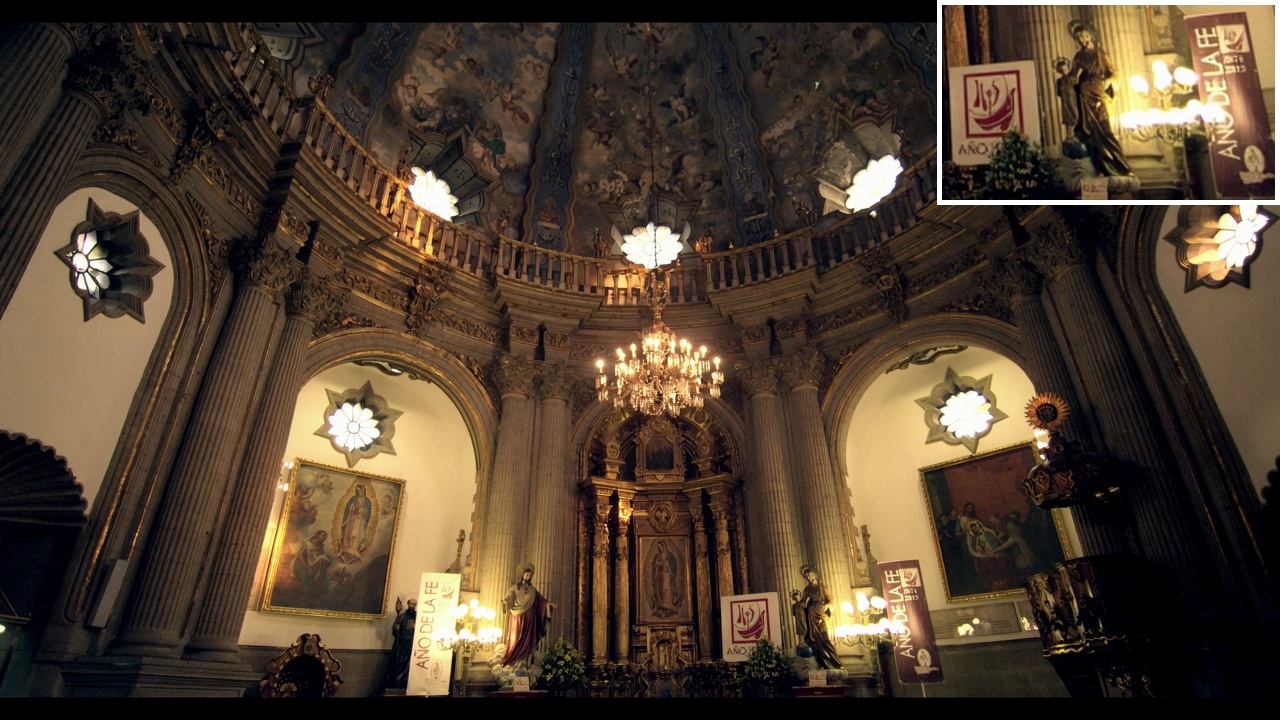}
    \caption{48$\times$ Hunyuan VAE}
    \label{fig:vae_vis_hunyuan}
  \end{subfigure}
  \hfill
  
  \caption{VAE visualization comparison at 30 fps, with resolution of 720$\times$1280. Better view with zoom-in. }
  \label{fig:vae_qualitative_comparison}
\end{figure}

We also present qualitative results of our VAE compared to the best baseline in \Cref{fig:vae_qualitative_comparison}. Our VAE provides competitive reconstruction details and fidelity for example when compared to HunyuanVideo, even with a higher compression ratio. For more visualization, please refer to our website.

%% file: sections/5_applications.tex
\section{Applications}
\label{sec:applications}

A hallmark characteristic of the foundation model is to support a diverse range of downstream video tasks, either through zero-shot generation, or lightweight (\eg, LoRA) fine-tuning. To demonstrate this capability, we conduct a qualitative study on the video applications built on the \modelname{} foundation model. We briefly discuss these tasks. Note that some of these works have been published; we refer readers to the corresponding papers or the website for the generated videos.

\paragraph{Image-to-video generation.}
Our model is trained with both text-to-video and image-to-video objectives, enabling it to natively generate video from an image and a text prompt. By conditioning on the first and last frames, it can also perform video transitions between the two input frames.
Additionally, by conditioning on the first and last frames, it can generate videos.

\paragraph{Human video generation.} Considering one significant domain of content generation is human generation, OmniHuman-1~\citep{lin2025omnihuman} leverages the generation fidelity and appealing quality of Seaweed and delivers state-of-the-art human animation models by modification on architecture, tailored training strategies and data. 

\paragraph{Subject-consistent video generation.} To make the generated concept controllable and aligned with human intent, our model can be finetuned to enable the generation of single/multiple subjects (\eg, identity of the reference face, objects, clothing, animals, virtual characters)~\citep{liu2025phantom}, allowing realistic interactions between multiple subjects, such as group interactions, product demonstrations, virtual try-on.

\begin{figure}
    \centering
    \includegraphics[width=0.75\linewidth]{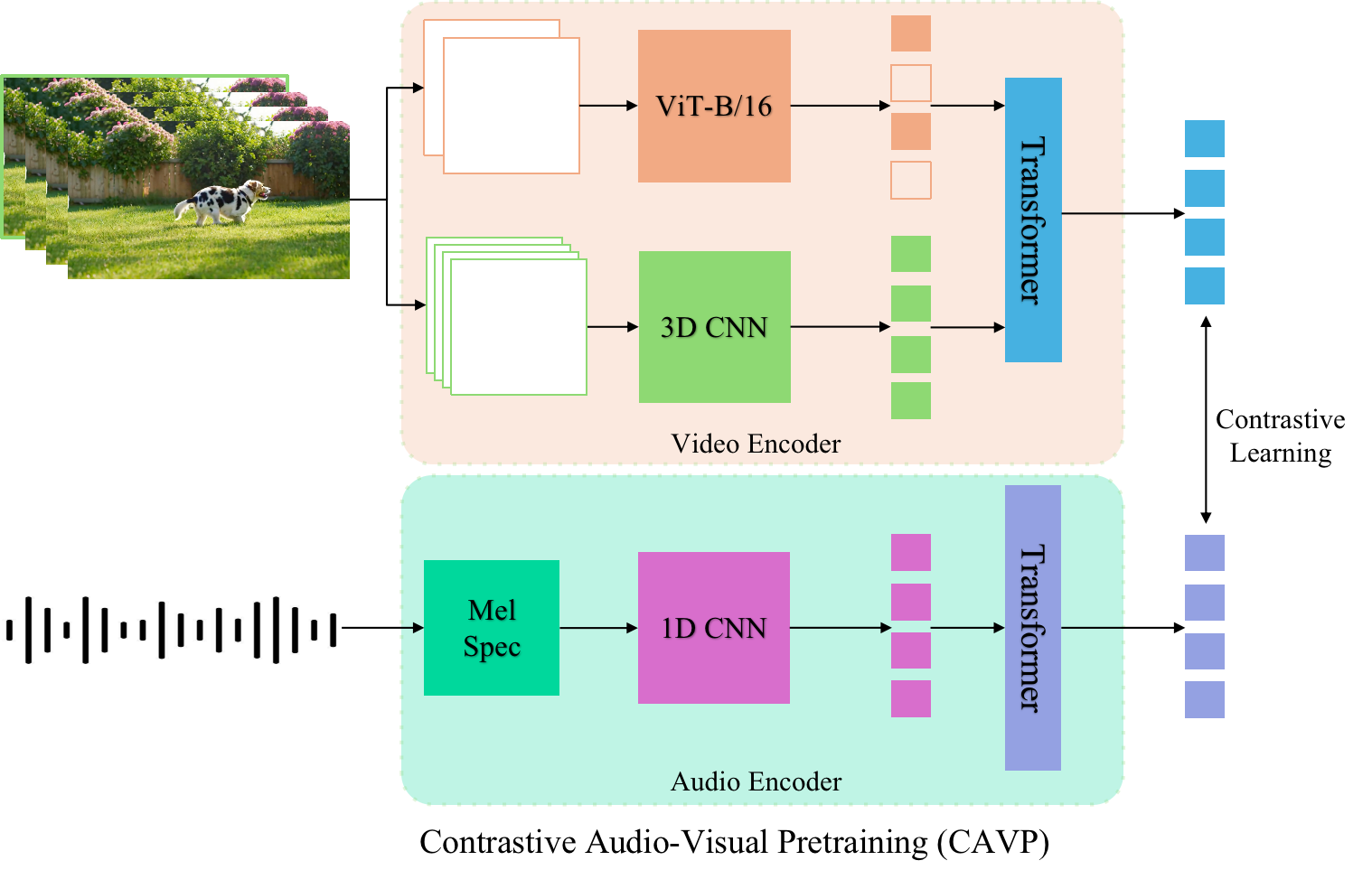}
    \vspace{-3mm}
    \caption{Contrastive Audio-Visual Pretraining (CAVP).}
    \label{fig:audio-cavp}
    \vspace{-5mm}
\end{figure}

\paragraph{Video-audio joint generation.}
We also design an audio generation model designed for producing high-quality audiovisual content. Instead of textual prompts, we condition the audio generation on video inputs to enhance cross-modal understanding and ensure temporal coherence across scenes. Inspired by image-text contrastive pre-training, fine-grained action semantics and temporal alignment can be effectively captured from audiovisual data.
Our Contrastive Audio-Visual Pretraining (CAVP) model features a dual-branch architecture for extracting video embeddings used in audio generation. As illustrated in \cref{fig:audio-cavp}, it includes a dual-branch video encoder: one branch uses higher FPS with a 3D CNN to extract fine-grained video embeddings, while the other operates on keyframes at lower FPS with a pre-trained SigLIP model. The audio encoder employs mel-spectrogram-based 1D CNNs and a transformer network to extract audio features. The video encoder, trained contrastively, serves as the condition for audio generation.
For audio generation, we utilize a conditional latent diffusion model with a flow-matching objective. The video embedding is temporally upsampled to match the audio FPS and concatenated with global musical embeddings and frame-level speech embeddings to serve as conditioning inputs for the audio generation process.

\paragraph{Long-video generation and storytelling.} Since currently video generator produces single-shot videos last 5-10 seconds, long-context tuning (LCT)~\citep{guo2025long} is proposed to adapt single-shot Seaweed for scene-level generation. LCT enables the auto-regressive rollout and exhibits emerging capabilities like composable generation and shot extension. To generate the story script, we also explore interleaved video and text generation in the context of video narrative creation, as proposed in~\cite{xiao2025videoauteur}.

\paragraph{Real-time generation.} Diffusion models require multiple steps for denoising videos. This is time-consuming and limits various applications. Seaweed-APT~\citep{lin2025diffusion} proposes adversarial post-training to enable one-step generation. It is the first approach to demonstrate real-time video synthesis at 1280$\times$720 resolution and 24fps, unlocking a wide range of real-time applications.

\paragraph{Super-resolution generation.} Seaweed can not only produce high-resolution videos directly but also serves as a starting point for arbitrary length and resolution video restoration~\citep{wang2025seedvr} that achieves highly the state-of-the-art performance on public benchmarks, as well as AI-generated videos.

\paragraph{Camera controlled generation.} The camera plays an important role in video generation. Our model supports various camera categories (\eg, Dolly-out, Pan-left, Spin-shot) using the synthetic data~\cite{zhao2025synthetic}. We also introduce CameraCtrl II~\citep{he2025cameractrl} for precise controllability. Together with image-to-video, long-video, and real-time generation, CameraCtrl II allows users to explore the generated world.

%% file: sections/6_related_work.tex
\section{Related Work}

Video generation has seen rapid progress over the past two years, emerging as a key area of interest in multimodal AI research. Its potential has long been recognized due to its ability to integrate various modalities, \ie text, image, audio, and video, into a unified generative framework. This capability enables a wide range of applications, including text-to-video, image-to-video, and audio-driven video synthesis.

While many techniques are inspired by those in image generation, video generation faces unique challenges. Unlike static images, videos require the modeling of motion dynamics and maintaining temporal coherence across long sequences. These demands tremendously increase the computational complexity of both training and inference, making video generation models one of the most resource-intensive foundation models to develop and deploy.

In the diffusion model era, this computational burden initially led to strategies that fine-tuned pre-trained text-to-image models. For instance, researchers extended U-Net backbones~\cite{ronneberger2015u} with temporal layers~\cite{singer2022make,blattmann2023align} or modified existing architectures~\cite{meng2021sdedit,esser2023structure,liew2023magicedit} to accommodate video data. Meanwhile, approaches trained models from scratch are proposed to demonstrate the benefit of jointly learning from both images and videos~\cite{ho2022video,gupta2024photorealistic}. This approach, now widely adopted, helps models generalize while leveraging large-scale datasets. Several architectural innovations emerged during this period. Diffusion transformers (DiTs)~\cite{peebles2023scalable} presented a scalable alternative to convolutional U-Nets, offering improved modeling capacity.
In parallel, causal 3D variational autoencoders (VAEs)~\cite{yu2023magvit,yu2023language} were introduced as an advancement over traditional 2D VAEs, which offers improved modeling of spatiotemporal representations for video. These components have since become standard techniques in modern video generation systems. WALT exemplifies this shift, employing a diffusion transformer trained from scratch in combination with MAGVIT, a 3D VAE~\cite{yu2023language}.

The introduction of Sora~\cite{sora} in early 2024 marked a significant milestone in video generation. It combined many of the most effective prior techniques such as DiTs, 3D VAEs, joint image-video training, and sequence packing~\cite{dehghani2024patch}, into a unified, scalable training system. By scaling both model size and dataset scale, Sora achieved video generation quality that significantly surpassed that of previous models. In addition, Sora introduced novel techniques, including training at native resolutions and enabling multi-resolution generation within transformer architectures.

Sora’s impressive demonstrations ignited widespread interest across both academia and industry, fueling increased investment in GPU infrastructure and research dedicated to building foundation models for video generation. Training these models typically demands substantial computational resources—often requiring thousands of GPUs.
Consequently, the development of such models is largely concentrated within industrial research labs, with many released as commercial products in response to the growing market demand for high-quality video generation. Examples include MovieGen~\cite{polyak2025moviegencastmedia}, HunyuanVideo\cite{kong2024hunyuanvideo}, Nova~\cite{intelligence2024amazon}, Cosmos~\cite{nvidia2025cosmosworldfoundationmodel}, Veo~\cite{veo2}, Pika~\cite{pika}, Runway~\cite{runway}, Kling~\cite{kling}, WanVideo~\cite{wan2.1}, Pixelverse~\cite{pixelverse}, Hailuo~\cite{hailuo}, Mochi~\cite{mochi}, Vidu~\cite{vidu}, and DreamMachine~\cite{dreammachine}, among others. Some models are accompanied by public technical reports that discuss their detailed design \cite{polyak2025moviegencastmedia,nvidia2025cosmosworldfoundationmodel,kong2024hunyuanvideo}. In this technical report, we present a resource-efficient video generation model that achieves competitive performance with significantly lower computational cost. We detail the key architectural and training design choices that enable this efficiency, aiming to contribute practical insights that complement existing approaches in the literature.

%% file: sections/7_conclusion.tex
\section{Conclusion}

We present a cost-effective video generation foundation model with 7 billion parameters. Our findings show that despite using moderate computational resources, \modelname{} matches or exceeds the performance of larger models trained with significantly more GPU resources, demonstrating strong generalization across diverse video generation tasks. These results verify our discussed design choices and highlight the potential of medium-sized models as efficient video foundation models. We hope our insights into model design and training strategies will inspire further advancements in video generation research.

Despite its promising capabilities, our model has several known limitations. First, there remains significant room for improvement across nearly all aspects of video foundation models. Addressing these challenges will require a collaborative effort from both the research community and industry to drive advancements in data curation, model design, and post-training. Second, due to limited computational capacity, our model exhibits limitations in generating fine-grained details, such as small faces or delicate patterns. Finally, ensuring responsible video generation remains a critical area of research. More efforts are needed to enhance safety, fairness, and ethical considerations in video generation.

\section*{Contributors and Acknowledgments}

Below are the researchers (individuals who worked full-time on the project) and contributors (individuals who provided assistance to the project). $*$ indicates that the list is ordered alphabetically. The Points of Contact (POCs) in the infrastructure work stream are marked with $\dagger$.

\subsection*{Research Team}

\textbf{Model}
\begin{itemize}
    \item Code \& Engineering: Shanchuan Lin, Peihao Zhu, Qi Zhao.
    \item Modeling Call: Ceyuan Yang.
    \item VAE: Yang Zhao, Hao Chen.
    \item DiT: Zhijie Lin, Fei Xiao.
    \item Post-training$^*$: Feng Cheng, Haoyuan Guo, Meng Wei, Zhiwu Qing.
\end{itemize}







\textbf{Data}
\begin{itemize}
    \item Processing Algorithms$^*$: Fangyuan Kong, Jiangqiao Yan, Liangke Gui, Lu Qi, Sen Wang, Tuyen Hoang, Ziyan Yang.
    \item Processing Pipeline: Zhibei Ma, Sheng Bi.
    \item Acquisition: Feilong Zuo, Siyu Zhang.
    \item Evaluation \& Annotation: Xuejiao Zeng.
\end{itemize}





\textbf{Research Lead:}
Lu Jiang, Jiashi Feng, Zhenheng Yang, Jianchao Yang.

\textbf{Infrastructure$^*$:}
Feng Ling, 
Heng Zhang, 
Houmin Wei, 
Huafeng Kuang, 
Huixia Li$^\dagger$, 
Jerry Duncan, 
Jiashi Li$^\dagger$, 
Junda Zhang, 
Junru Zheng, 
Li Sun, 
Manlin Zhang,
Renfei Sun,
Rui Wang$^\dagger$, 
Shu Liu$^\dagger$, 
Xiaojie Li,
Xin Xia,
Xuefeng Xiao$^\dagger$, 
Xuyan Chi, 
Yanghua Peng, 
Yuxi Ren$^\dagger$, 
Zhongkai Zhao, 
Zuquan Song.

\textbf{Contributors$^*$:} Bingchuan Li, Chao Liang, Deyao Zhu, Gaojie Lin, Gen Li, Hao He, Jianwen Jiang, Jianyi Wang, Jiaqi Yang, Jiawei Liu, Junfei Xiao, Lijie Liu, Lizhen Wang, Longhao Zhang, Qian He, Ruiqi Xia, Siyu Zhou, Tianshu Hu, Tianxiang Ma, Weilin Huang, Xiaobin Zhuang, Xiaohui Shen, Xinglong Wu, Yongming Zhu, Yuping Wang, Yuwei Guo, Yuxuan Luo, Yuxuan Wang, Zerong Zheng, Zhengkun Rong, Zhuo Chen, Zhuowei Chen.

\subsection*{Acknowledgment}
We would like to thank Agnes Li, Tim Wang, Yi Luo, and Zhang Xiaoyi for providing support. Special thanks to the leadership team: Wenjia Zhu and Yonghui Wu for their valuable research discussions and support.

%% file: main.bbl
\begin{thebibliography}{97}
\providecommand{\natexlab}[1]{#1}
\providecommand{\url}[1]{\texttt{#1}}
\expandafter\ifx\csname urlstyle\endcsname\relax
  \providecommand{\doi}[1]{doi: #1}\else
  \providecommand{\doi}{doi: \begingroup \urlstyle{rm}\Url}\fi

\bibitem[{Adobe Blog}(2024)]{firefly}
{Adobe Blog}.
\newblock Bringing generative ai to video with adobe firefly video model.
\newblock \url{https://blog.adobe.com/en/publish/2024/09/11/bringing-gen-ai-to-video-adobe-firefly-video-model-coming-soon}, 2024.

\bibitem[Agarwal et~al.(2025{\natexlab{a}})Agarwal, Ali, Bala, Balaji, Barker, Cai, Chattopadhyay, Chen, Cui, Ding, Dworakowski, Fan, Fenzi, Ferroni, Fidler, et~al.]{nvidia2025cosmosworldfoundationmodel}
Niket Agarwal, Arslan Ali, Maciej Bala, Yogesh Balaji, Erik Barker, Tiffany Cai, Prithvijit Chattopadhyay, Yongxin Chen, Yin Cui, Yifan Ding, Daniel Dworakowski, Jiaojiao Fan, Michele Fenzi, Francesco Ferroni, Sanja Fidler, et~al.
\newblock Cosmos world foundation model platform for physical ai, 2025{\natexlab{a}}.

\bibitem[Agarwal et~al.(2025{\natexlab{b}})Agarwal, Ali, Bala, Balaji, Barker, Cai, Chattopadhyay, Chen, Cui, Ding, et~al.]{agarwal2025cosmos}
Niket Agarwal, Arslan Ali, Maciej Bala, Yogesh Balaji, Erik Barker, Tiffany Cai, Prithvijit Chattopadhyay, Yongxin Chen, Yin Cui, Yifan Ding, et~al.
\newblock Cosmos world foundation model platform for physical ai.
\newblock \emph{arXiv preprint arXiv:2501.03575}, 2025{\natexlab{b}}.

\bibitem[Alayrac et~al.(2022)Alayrac, Donahue, Luc, Miech, Barr, Hasson, Lenc, Mensch, Millican, Reynolds, Ring, Rutherford, Cabi, Han, Gong, Samangooei, Monteiro, Menick, Borgeaud, Brock, Nematzadeh, Sharifzadeh, Binkowski, Barreira, Vinyals, Zisserman, and Simonyan]{alayrac2022flamingo}
Jean-Baptiste Alayrac, Jeff Donahue, Pauline Luc, Antoine Miech, Iain Barr, Yana Hasson, Karel Lenc, Arthur Mensch, Katie Millican, Malcolm Reynolds, Roman Ring, Eliza Rutherford, Serkan Cabi, Tengda Han, Zhitao Gong, Sina Samangooei, Mariana Monteiro, Jacob Menick, Sebastian Borgeaud, Andy Brock, Aida Nematzadeh, Sahand Sharifzadeh, Mikolaj Binkowski, Rodrigo Barreira, Oriol Vinyals, Andrew Zisserman, and Karen Simonyan.
\newblock Flamingo: A visual language model for few-shot learning.
\newblock In \emph{NeurIPS}, 2022.

\bibitem[Blattmann et~al.(2023{\natexlab{a}})Blattmann, Dockhorn, Kulal, Mendelevitch, Kilian, Lorenz, Levi, English, Voleti, Letts, Jampani, and Rombach]{blattmann2023stablevideodiffusionscaling}
Andreas Blattmann, Tim Dockhorn, Sumith Kulal, Daniel Mendelevitch, Maciej Kilian, Dominik Lorenz, Yam Levi, Zion English, Vikram Voleti, Adam Letts, Varun Jampani, and Robin Rombach.
\newblock Stable video diffusion: Scaling latent video diffusion models to large datasets, 2023{\natexlab{a}}.

\bibitem[Blattmann et~al.(2023{\natexlab{b}})Blattmann, Rombach, Ling, Dockhorn, Kim, Fidler, and Kreis]{blattmann2023align}
Andreas Blattmann, Robin Rombach, Huan Ling, Tim Dockhorn, Seung~Wook Kim, Sanja Fidler, and Karsten Kreis.
\newblock Align your latents: High-resolution video synthesis with latent diffusion models.
\newblock In \emph{CVPR}, pages 22563--22575, 2023{\natexlab{b}}.

\bibitem[Bommasani et~al.(2021)Bommasani, Hudson, Adeli, Altman, Arora, von Arx, Bernstein, Bohg, Bosselut, Brunskill, et~al.]{bommasani2021opportunities}
Rishi Bommasani, Drew~A Hudson, Ehsan Adeli, Russ Altman, Simran Arora, Sydney von Arx, Michael~S Bernstein, Jeannette Bohg, Antoine Bosselut, Emma Brunskill, et~al.
\newblock On the opportunities and risks of foundation models.
\newblock \emph{arXiv preprint arXiv:2108.07258}, 2021.

\bibitem[Borsos et~al.(2023)Borsos, Marinier, Vincent, Kharitonov, Pietquin, Sharifi, Roblek, Teboul, Grangier, Tagliasacchi, et~al.]{borsos2023audiolm}
Zal{\'a}n Borsos, Rapha{\"e}l Marinier, Damien Vincent, Eugene Kharitonov, Olivier Pietquin, Matt Sharifi, Dominik Roblek, Olivier Teboul, David Grangier, Marco Tagliasacchi, et~al.
\newblock Audiolm: a language modeling approach to audio generation.
\newblock \emph{IEEE/ACM TASLP}, 31:\penalty0 2523--2533, 2023.

\bibitem[Brown et~al.(2020)Brown, Mann, Ryder, Subbiah, Kaplan, Dhariwal, Neelakantan, Shyam, Sastry, Askell, et~al.]{brown2020language}
Tom Brown, Benjamin Mann, Nick Ryder, Melanie Subbiah, Jared~D Kaplan, Prafulla Dhariwal, Arvind Neelakantan, Pranav Shyam, Girish Sastry, Amanda Askell, et~al.
\newblock Language models are few-shot learners.
\newblock \emph{NeurIPS}, 33:\penalty0 1877--1901, 2020.

\bibitem[{ByteDance}(2024)]{bmf}
{ByteDance}.
\newblock bmf.
\newblock \url{https://babitmf.github.io/}, 2024.

\bibitem[Chen et~al.(2023)Chen, Yu, Ge, Yao, Xie, Wu, Wang, Kwok, Luo, Lu, et~al.]{chen2023pixart}
Junsong Chen, Jincheng Yu, Chongjian Ge, Lewei Yao, Enze Xie, Yue Wu, Zhongdao Wang, James Kwok, Ping Luo, Huchuan Lu, et~al.
\newblock Pixart-alpha: Fast training of diffusion transformer for photorealistic text-to-image synthesis.
\newblock \emph{arXiv preprint arXiv:2310.00426}, 2023.

\bibitem[Chen et~al.(2016)Chen, Xu, Zhang, and Guestrin]{chen2016training}
Tianqi Chen, Bing Xu, Chiyuan Zhang, and Carlos Guestrin.
\newblock Training deep nets with sublinear memory cost.
\newblock \emph{arXiv preprint arXiv:1604.06174}, 2016.

\bibitem[Chen et~al.(2025)Chen, Liu, Chen, Feng, Liu, Shen, and Zhao]{chen2025livephoto}
Xi~Chen, Zhiheng Liu, Mengting Chen, Yutong Feng, Yu~Liu, Yujun Shen, and Hengshuang Zhao.
\newblock Livephoto: Real image animation with text-guided motion control.
\newblock In \emph{ECCV}, 2025.

\bibitem[Chowdhery et~al.(2023)Chowdhery, Narang, Devlin, Bosma, Mishra, Roberts, Barham, Chung, Sutton, Gehrmann, et~al.]{chowdhery2023palm}
Aakanksha Chowdhery, Sharan Narang, Jacob Devlin, Maarten Bosma, Gaurav Mishra, Adam Roberts, Paul Barham, Hyung~Won Chung, Charles Sutton, Sebastian Gehrmann, et~al.
\newblock Palm: Scaling language modeling with pathways.
\newblock \emph{JMLR}, 24\penalty0 (240):\penalty0 1--113, 2023.

\bibitem[Dehghani et~al.(2024)Dehghani, Mustafa, Djolonga, Heek, Minderer, Caron, Steiner, Puigcerver, Geirhos, Alabdulmohsin, et~al.]{dehghani2024patch}
Mostafa Dehghani, Basil Mustafa, Josip Djolonga, Jonathan Heek, Matthias Minderer, Mathilde Caron, Andreas Steiner, Joan Puigcerver, Robert Geirhos, Ibrahim~M Alabdulmohsin, et~al.
\newblock Patch n’pack: Navit, a vision transformer for any aspect ratio and resolution.
\newblock \emph{NeurIPS}, 36, 2024.

\bibitem[Developers({\natexlab{a}})]{ffmpeg}
FFmpeg Developers.
\newblock Ffmpeg, {\natexlab{a}}.
\newblock \url{https://ffmpeg.org/}.

\bibitem[Developers({\natexlab{b}})]{pyscenedetect}
PySceneDetect Developers.
\newblock Pyscenedetect, {\natexlab{b}}.
\newblock \url{https://www.scenedetect.com/}.

\bibitem[Elo and Sloan(2008)]{elo1978rating}
Arpad~E. Elo and Sam Sloan.
\newblock \emph{The Rating of Chess Players, Past and Present}.
\newblock Ishi Press, 2008.

\bibitem[{Epic Games}(2024)]{UE5}
{Epic Games}.
\newblock Unreal engine 5.
\newblock \url{https://www.unrealengine.com/en-US/unreal-engine-5}, 2024.

\bibitem[Esser et~al.(2023)Esser, Chiu, Atighehchian, Granskog, and Germanidis]{esser2023structure}
Patrick Esser, Johnathan Chiu, Parmida Atighehchian, Jonathan Granskog, and Anastasis Germanidis.
\newblock Structure and content-guided video synthesis with diffusion models.
\newblock In \emph{CVPR}, pages 7346--7356, 2023.

\bibitem[Esser et~al.(2024)Esser, Kulal, Blattmann, Entezari, M{\"u}ller, Saini, Levi, Lorenz, Sauer, Boesel, et~al.]{esser2024scaling}
Patrick Esser, Sumith Kulal, Andreas Blattmann, Rahim Entezari, Jonas M{\"u}ller, Harry Saini, Yam Levi, Dominik Lorenz, Axel Sauer, Frederic Boesel, et~al.
\newblock Scaling rectified flow transformers for high-resolution image synthesis.
\newblock In \emph{ICML}, 2024.

\bibitem[{Genmoai}(2024)]{mochi}
{Genmoai}.
\newblock Mochi.
\newblock \url{https://github.com/genmoai/mochi}, 2024.

\bibitem[Girdhar et~al.(2023)Girdhar, El-Nouby, Liu, Singh, Alwala, Joulin, and Misra]{girdhar2023imagebind}
Rohit Girdhar, Alaaeldin El-Nouby, Zhuang Liu, Mannat Singh, Kalyan~Vasudev Alwala, Armand Joulin, and Ishan Misra.
\newblock Imagebind: One embedding space to bind them all.
\newblock In \emph{CVPR}, 2023.

\bibitem[{Google}(2024)]{veo2}
{Google}.
\newblock Veo-2.
\newblock \url{https://deepmind.google/technologies/veo/veo-2/}, 2024.

\bibitem[Grattafiori et~al.(2024)Grattafiori, Dubey, Jauhri, Pandey, Kadian, Al-Dahle, Letman, Mathur, Schelten, Vaughan, et~al.]{grattafiori2024llama}
Aaron Grattafiori, Abhimanyu Dubey, Abhinav Jauhri, Abhinav Pandey, Abhishek Kadian, Ahmad Al-Dahle, Aiesha Letman, Akhil Mathur, Alan Schelten, Alex Vaughan, et~al.
\newblock The llama 3 herd of models.
\newblock \emph{arXiv preprint arXiv:2407.21783}, 2024.

\bibitem[Guo et~al.(2025)Guo, Yang, Yang, Ma, Lin, Yang, Lin, and Jiang]{guo2025long}
Yuwei Guo, Ceyuan Yang, Ziyan Yang, Zhibei Ma, Zhijie Lin, Zhenheng Yang, Dahua Lin, and Lu~Jiang.
\newblock Long context tuning for video generation.
\newblock \emph{arXiv preprint arXiv:2503.10589}, 2025.

\bibitem[Gupta et~al.(2024)Gupta, Yu, Sohn, Gu, Hahn, Li, Essa, Jiang, and Lezama]{gupta2024photorealistic}
Agrim Gupta, Lijun Yu, Kihyuk Sohn, Xiuye Gu, Meera Hahn, Fei-Fei Li, Irfan Essa, Lu~Jiang, and Jos{\'e} Lezama.
\newblock Photorealistic video generation with diffusion models.
\newblock In \emph{European Conference on Computer Vision}, pages 393--411. Springer, 2024.

\bibitem[HaCohen et~al.(2024)HaCohen, Chiprut, Brazowski, Shalem, Moshe, Richardson, Levin, Shiran, Zabari, Gordon, Panet, Weissbuch, Kulikov, Bitterman, Melumian, and Bibi]{HaCohen2024LTXVideo}
Yoav HaCohen, Nisan Chiprut, Benny Brazowski, Daniel Shalem, Dudu Moshe, Eitan Richardson, Eran Levin, Guy Shiran, Nir Zabari, Ori Gordon, Poriya Panet, Sapir Weissbuch, Victor Kulikov, Yaki Bitterman, Zeev Melumian, and Ofir Bibi.
\newblock Ltx-video: Realtime video latent diffusion.
\newblock \emph{arXiv preprint arXiv:2501.00103}, 2024.

\bibitem[He et~al.(2025)He, Yang, Lin, Xu, Wei, Gui, Zhao, Wetzstein, Jiang, and Li]{he2025cameractrl}
Hao He, Ceyuan Yang, Shanchuan Lin, Yinghao Xu, Meng Wei, Liangke Gui, Qi~Zhao, Gordon Wetzstein, Lu~Jiang, and Hongsheng Li.
\newblock Cameractrl ii: Dynamic scene exploration via camera-controlled video diffusion models.
\newblock \emph{arXiv preprint arXiv:2503.10592}, 2025.

\bibitem[Ho and Salimans(2022)]{ho2022classifier}
Jonathan Ho and Tim Salimans.
\newblock Classifier-free diffusion guidance.
\newblock \emph{arXiv preprint arXiv:2207.12598}, 2022.

\bibitem[Ho et~al.(2022)Ho, Salimans, Gritsenko, Chan, Norouzi, and Fleet]{ho2022video}
Jonathan Ho, Tim Salimans, Alexey Gritsenko, William Chan, Mohammad Norouzi, and David~J Fleet.
\newblock Video diffusion models.
\newblock \emph{arXiv:2204.03458}, 2022.

\bibitem[Intelligence(2024)]{intelligence2024amazon}
Amazon Artificial~General Intelligence.
\newblock The amazon nova family of models: Technical report and model card.
\newblock 2024.

\bibitem[Ioffe and Szegedy(2015)]{ioffe2015batchnormalizationacceleratingdeep}
Sergey Ioffe and Christian Szegedy.
\newblock Batch normalization: Accelerating deep network training by reducing internal covariate shift, 2015.
\newblock ICML.

\bibitem[Isola et~al.(2018)Isola, Zhu, Zhou, and Efros]{isola2018imagetoimagetranslationconditionaladversarial}
Phillip Isola, Jun-Yan Zhu, Tinghui Zhou, and Alexei~A. Efros.
\newblock Image-to-image translation with conditional adversarial networks, 2018.

\bibitem[Jacobs et~al.(2023)Jacobs, Tanaka, Zhang, Zhang, Song, Rajbhandari, and He]{jacobs2023deepspeed}
Sam~Ade Jacobs, Masahiro Tanaka, Chengming Zhang, Minjia Zhang, Shuaiwen~Leon Song, Samyam Rajbhandari, and Yuxiong He.
\newblock Deepspeed ulysses: System optimizations for enabling training of extreme long sequence transformer models.
\newblock \emph{arXiv preprint arXiv:2309.14509}, 2023.

\bibitem[Jiang et~al.(2023)Jiang, Sablayrolles, Mensch, Bamford, Chaplot, de~las Casas, Bressand, Lengyel, Lample, Saulnier, Lavaud, Lachaux, Stock, Scao, Lavril, Wang, Lacroix, and Sayed]{jiang2023mistral7b}
Albert~Q. Jiang, Alexandre Sablayrolles, Arthur Mensch, Chris Bamford, Devendra~Singh Chaplot, Diego de~las Casas, Florian Bressand, Gianna Lengyel, Guillaume Lample, Lucile Saulnier, Lélio~Renard Lavaud, Marie-Anne Lachaux, Pierre Stock, Teven~Le Scao, Thibaut Lavril, Thomas Wang, Timothée Lacroix, and William~El Sayed.
\newblock Mistral 7b.
\newblock \emph{arXiv preprint arXiv:2310.06825}, 2023.

\bibitem[Karras et~al.(2020)Karras, Laine, Aittala, Hellsten, Lehtinen, and Aila]{karras2020analyzingimprovingimagequality}
Tero Karras, Samuli Laine, Miika Aittala, Janne Hellsten, Jaakko Lehtinen, and Timo Aila.
\newblock Analyzing and improving the image quality of stylegan, 2020.

\bibitem[Kingma(2013)]{kingma2013auto}
Diederik~P Kingma.
\newblock Auto-encoding variational bayes.
\newblock \emph{arXiv preprint arXiv:1312.6114}, 2013.

\bibitem[Kondratyuk et~al.(2023)Kondratyuk, Yu, Gu, Lezama, Huang, Schindler, Hornung, Birodkar, Yan, Chiu, et~al.]{kondratyuk2023videopoet}
Dan Kondratyuk, Lijun Yu, Xiuye Gu, Jos{\'e} Lezama, Jonathan Huang, Grant Schindler, Rachel Hornung, Vighnesh Birodkar, Jimmy Yan, Ming-Chang Chiu, et~al.
\newblock Videopoet: A large language model for zero-shot video generation.
\newblock \emph{arXiv preprint arXiv:2312.14125}, 2023.

\bibitem[Kong et~al.(2024)Kong, Tian, Zhang, Min, Dai, Zhou, Xiong, Li, Wu, Zhang, et~al.]{kong2024hunyuanvideo}
Weijie Kong, Qi~Tian, Zijian Zhang, Rox Min, Zuozhuo Dai, Jin Zhou, Jiangfeng Xiong, Xin Li, Bo~Wu, Jianwei Zhang, et~al.
\newblock Hunyuanvideo: A systematic framework for large video generative models.
\newblock \emph{arXiv preprint arXiv:2412.03603}, 2024.

\bibitem[{Kuaishou}(2024)]{kling}
{Kuaishou}.
\newblock Kling video model.
\newblock \url{https://kling.kuaishou.com/en}, 2024.

\bibitem[Li et~al.(2024)Li, Lin, Ye, Chen, Cheng, Yuan, and Yuan]{li2024wf}
Zongjian Li, Bin Lin, Yang Ye, Liuhan Chen, Xinhua Cheng, Shenghai Yuan, and Li~Yuan.
\newblock Wf-vae: Enhancing video vae by wavelet-driven energy flow for latent video diffusion model.
\newblock \emph{arXiv preprint arXiv:2411.17459}, 2024.

\bibitem[Liang et~al.(2024)Liang, He, Yang, and Dai]{liang2024scaling}
Zhengyang Liang, Hao He, Ceyuan Yang, and Bo~Dai.
\newblock Scaling laws for diffusion transformers.
\newblock \emph{arXiv preprint arXiv:2410.08184}, 2024.

\bibitem[Liew et~al.(2023)Liew, Yan, Zhang, Xu, and Feng]{liew2023magicedit}
Jun~Hao Liew, Hanshu Yan, Jianfeng Zhang, Zhongcong Xu, and Jiashi Feng.
\newblock Magicedit: High-fidelity and temporally coherent video editing.
\newblock \emph{arXiv preprint arXiv:2308.14749}, 2023.

\bibitem[Lin et~al.(2024)Lin, Ge, Cheng, Li, Zhu, Wang, He, Ye, Yuan, Chen, et~al.]{lin2024open}
Bin Lin, Yunyang Ge, Xinhua Cheng, Zongjian Li, Bin Zhu, Shaodong Wang, Xianyi He, Yang Ye, Shenghai Yuan, Liuhan Chen, et~al.
\newblock Open-sora plan: Open-source large video generation model.
\newblock \emph{arXiv preprint arXiv:2412.00131}, 2024.

\bibitem[Lin et~al.(2025{\natexlab{a}})Lin, Jiang, Yang, Zheng, and Liang]{lin2025omnihuman}
Gaojie Lin, Jianwen Jiang, Jiaqi Yang, Zerong Zheng, and Chao Liang.
\newblock Omnihuman-1: Rethinking the scaling-up of one-stage conditioned human animation models.
\newblock \emph{arXiv preprint arXiv:2502.01061}, 2025{\natexlab{a}}.

\bibitem[Lin et~al.(2025{\natexlab{b}})Lin, Xia, Ren, Yang, Xiao, and Jiang]{lin2025diffusion}
Shanchuan Lin, Xin Xia, Yuxi Ren, Ceyuan Yang, Xuefeng Xiao, and Lu~Jiang.
\newblock Diffusion adversarial post-training for one-step video generation.
\newblock \emph{arXiv preprint arXiv:2501.08316}, 2025{\natexlab{b}}.

\bibitem[Liu et~al.(2024{\natexlab{a}})Liu, Feng, Xue, Wang, Wu, Lu, Zhao, Deng, Zhang, Ruan, et~al.]{liu2024deepseek}
Aixin Liu, Bei Feng, Bing Xue, Bingxuan Wang, Bochao Wu, Chengda Lu, Chenggang Zhao, Chengqi Deng, Chenyu Zhang, Chong Ruan, et~al.
\newblock Deepseek-v3 technical report.
\newblock \emph{arXiv preprint arXiv:2412.19437}, 2024{\natexlab{a}}.

\bibitem[Liu et~al.(2024{\natexlab{b}})Liu, Li, Li, and Lee]{liu2024improved}
Haotian Liu, Chunyuan Li, Yuheng Li, and Yong~Jae Lee.
\newblock Improved baselines with visual instruction tuning.
\newblock In \emph{Proceedings of the IEEE/CVF Conference on Computer Vision and Pattern Recognition}, pages 26296--26306, 2024{\natexlab{b}}.

\bibitem[Liu et~al.(2025)Liu, Ma, Li, Chen, Liu, He, and Wu]{liu2025phantom}
Lijie Liu, Tianxaing Ma, Bingchuan Li, Zhuowei Chen, Jiawei Liu, Qian He, and Xinglong Wu.
\newblock Phantom: Subject-consistent video generation via cross-modal alignment.
\newblock \emph{arXiv preprint arXiv:2502.11079}, 2025.

\bibitem[Liu et~al.(2021)Liu, Lin, Cao, Hu, Wei, Zhang, Lin, and Guo]{liu2021swin}
Ze~Liu, Yutong Lin, Yue Cao, Han Hu, Yixuan Wei, Zheng Zhang, Stephen Lin, and Baining Guo.
\newblock Swin transformer: Hierarchical vision transformer using shifted windows.
\newblock In \emph{ICCV}, pages 10012--10022, 2021.

\bibitem[Liu et~al.(2022)Liu, Ning, Cao, Wei, Zhang, Lin, and Hu]{liu2022video}
Ze~Liu, Jia Ning, Yue Cao, Yixuan Wei, Zheng Zhang, Stephen Lin, and Han Hu.
\newblock Video swin transformer.
\newblock In \emph{CVPR}, pages 3202--3211, 2022.

\bibitem[{Luma}(2024)]{dreammachine}
{Luma}.
\newblock Dream machine.
\newblock \url{https://lumalabs.ai/dream-machine}, 2024.

\bibitem[Ma et~al.(2025)Ma, Huang, Yan, Chen, Duan, Yin, Wan, Ming, Song, Chen, et~al.]{ma2025step}
Guoqing Ma, Haoyang Huang, Kun Yan, Liangyu Chen, Nan Duan, Shengming Yin, Changyi Wan, Ranchen Ming, Xiaoniu Song, Xing Chen, et~al.
\newblock Step-video-t2v technical report: The practice, challenges, and future of video foundation model.
\newblock \emph{arXiv preprint arXiv:2502.10248}, 2025.

\bibitem[Meng et~al.(2021)Meng, He, Song, Song, Wu, Zhu, and Ermon]{meng2021sdedit}
Chenlin Meng, Yutong He, Yang Song, Jiaming Song, Jiajun Wu, Jun-Yan Zhu, and Stefano Ermon.
\newblock Sdedit: Guided image synthesis and editing with stochastic differential equations.
\newblock \emph{arXiv preprint arXiv:2108.01073}, 2021.

\bibitem[Mescheder et~al.(2018)Mescheder, Geiger, and Nowozin]{mescheder2018trainingmethodsgansactually}
Lars Mescheder, Andreas Geiger, and Sebastian Nowozin.
\newblock Which training methods for gans do actually converge?, 2018.

\bibitem[{Minimax}(2024)]{hailuo}
{Minimax}.
\newblock Hailuo.
\newblock \url{https://hailuoai.com/video}, 2024.

\bibitem[Miyato et~al.(2018)Miyato, Kataoka, Koyama, and Yoshida]{miyato2018spectralnormalizationgenerativeadversarial}
Takeru Miyato, Toshiki Kataoka, Masanori Koyama, and Yuichi Yoshida.
\newblock Spectral normalization for generative adversarial networks, 2018.

\bibitem[Moritz et~al.(2018)Moritz, Nishihara, Wang, Tumanov, Liaw, Liang, Elibol, Yang, Paul, Jordan, et~al.]{ray}
Philipp Moritz, Robert Nishihara, Stephanie Wang, Alexey Tumanov, Richard Liaw, Eric Liang, Melih Elibol, Zongheng Yang, William Paul, Michael~I Jordan, et~al.
\newblock Ray: A distributed framework for emerging $\{$AI$\}$ applications.
\newblock In \emph{13th USENIX symposium on operating systems design and implementation (OSDI 18)}, pages 561--577, 2018.

\bibitem[{OpenAI}(2024)]{sora}
{OpenAI}.
\newblock Sora.
\newblock \url{https://openai.com/sora/}, 2024.

\bibitem[Peebles and Xie(2023)]{peebles2023scalable}
William Peebles and Saining Xie.
\newblock Scalable diffusion models with transformers.
\newblock In \emph{ICCV}, pages 4195--4205, 2023.

\bibitem[{Pika}(2024)]{pika}
{Pika}.
\newblock Pika.
\newblock \url{https://pikartai.com/}, 2024.

\bibitem[{pixelverse}(2025)]{pixelverse}
{pixelverse}.
\newblock Pixelverse.
\newblock \url{https://www.pixelverse.xyz/}, 2025.

\bibitem[Polyak et~al.(2025)Polyak, Zohar, Brown, Tjandra, Sinha, Lee, Vyas, Shi, Ma, Chuang, Yan, et~al.]{polyak2025moviegencastmedia}
Adam Polyak, Amit Zohar, Andrew Brown, Andros Tjandra, Animesh Sinha, Ann Lee, Apoorv Vyas, Bowen Shi, Chih-Yao Ma, Ching-Yao Chuang, David Yan, et~al.
\newblock Movie gen: A cast of media foundation models, 2025.

\bibitem[Radford et~al.(2021)Radford, Kim, Hallacy, Ramesh, Goh, Agarwal, Sastry, Askell, Mishkin, Clark, et~al.]{radford2021learning}
Alec Radford, Jong~Wook Kim, Chris Hallacy, Aditya Ramesh, Gabriel Goh, Sandhini Agarwal, Girish Sastry, Amanda Askell, Pamela Mishkin, Jack Clark, et~al.
\newblock Learning transferable visual models from natural language supervision.
\newblock In \emph{ICML}, 2021.

\bibitem[Radford et~al.(2023)Radford, Kim, Xu, Brockman, McLeavey, and Sutskever]{radford2023robust}
Alec Radford, Jong~Wook Kim, Tao Xu, Greg Brockman, Christine McLeavey, and Ilya Sutskever.
\newblock Robust speech recognition via large-scale weak supervision.
\newblock In \emph{ICML}, pages 28492--28518. PMLR, 2023.

\bibitem[Rafailov et~al.(2023)Rafailov, Sharma, Mitchell, Manning, Ermon, and Finn]{rafailov2023direct}
Rafael Rafailov, Archit Sharma, Eric Mitchell, Christopher~D Manning, Stefano Ermon, and Chelsea Finn.
\newblock Direct preference optimization: Your language model is secretly a reward model.
\newblock \emph{NeurIPS}, 36:\penalty0 53728--53741, 2023.

\bibitem[Ren et~al.(2024)Ren, Xia, Lu, Zhang, Wu, Xie, Wang, and Xiao]{renhyper}
Yuxi Ren, Xin Xia, Yanzuo Lu, Jiacheng Zhang, Jie Wu, Pan Xie, Xing Wang, and Xuefeng Xiao.
\newblock Hyper-sd: Trajectory segmented consistency model for efficient image synthesis.
\newblock \emph{arXiv preprint arXiv:2404.13686}, 2024.

\bibitem[Rombach et~al.(2022)Rombach, Blattmann, Lorenz, Esser, and Ommer]{rombach2022high}
Robin Rombach, Andreas Blattmann, Dominik Lorenz, Patrick Esser, and Bj{\"o}rn Ommer.
\newblock High-resolution image synthesis with latent diffusion models.
\newblock In \emph{CVPR}, pages 10684--10695, 2022.

\bibitem[Ronneberger et~al.(2015)Ronneberger, Fischer, and Brox]{ronneberger2015u}
Olaf Ronneberger, Philipp Fischer, and Thomas Brox.
\newblock U-net: Convolutional networks for biomedical image segmentation.
\newblock In \emph{Medical image computing and computer-assisted intervention--MICCAI 2015: 18th international conference, Munich, Germany, October 5-9, 2015, proceedings, part III 18}, pages 234--241. Springer, 2015.

\bibitem[{Runway}(2024)]{runway}
{Runway}.
\newblock Runway-gen-3-alpha.
\newblock \url{https://runwayml.com/research/introducing-gen-3-alpha}, 2024.

\bibitem[Schönfeld et~al.(2021)Schönfeld, Schiele, and Khoreva]{schönfeld2021unetbaseddiscriminatorgenerative}
Edgar Schönfeld, Bernt Schiele, and Anna Khoreva.
\newblock A u-net based discriminator for generative adversarial networks, 2021.

\bibitem[Singer et~al.(2022)Singer, Polyak, Hayes, Yin, An, Zhang, Hu, Yang, Ashual, Gafni, et~al.]{singer2022make}
Uriel Singer, Adam Polyak, Thomas Hayes, Xi~Yin, Jie An, Songyang Zhang, Qiyuan Hu, Harry Yang, Oron Ashual, Oran Gafni, et~al.
\newblock Make-a-video: Text-to-video generation without text-video data.
\newblock \emph{arXiv preprint arXiv:2209.14792}, 2022.

\bibitem[Su et~al.(2024)Su, Ahmed, Lu, Pan, Bo, and Liu]{su2024roformer}
Jianlin Su, Murtadha Ahmed, Yu~Lu, Shengfeng Pan, Wen Bo, and Yunfeng Liu.
\newblock Roformer: Enhanced transformer with rotary position embedding.
\newblock \emph{Neurocomputing}, 568:\penalty0 127063, 2024.

\bibitem[Team(2025)]{wan2.1}
Wan Team.
\newblock Wan: Open and advanced large-scale video generative models.
\newblock 2025.

\bibitem[Teed and Deng(2020)]{teed2020raft}
Zachary Teed and Jia Deng.
\newblock Raft: Recurrent all-pairs field transforms for optical flow.
\newblock In \emph{ECCV}, 2020.

\bibitem[{The Blender Fundation}(2024)]{Blender}
{The Blender Fundation}.
\newblock Blender.
\newblock \url{https://www.blender.org/}, 2024.

\bibitem[Tseng et~al.(2021)Tseng, Jiang, Liu, Yang, and Yang]{lecamgan}
Hung-Yu Tseng, Lu~Jiang, Ce~Liu, Ming-Hsuan Yang, and Weilong Yang.
\newblock Regularing generative adversarial networks under limited data.
\newblock In \emph{CVPR}, 2021.

\bibitem[Unterthiner et~al.(2019)Unterthiner, van Steenkiste, Kurach, Marinier, Michalski, and Gelly]{unterthiner2019accurategenerativemodelsvideo}
Thomas Unterthiner, Sjoerd van Steenkiste, Karol Kurach, Raphael Marinier, Marcin Michalski, and Sylvain Gelly.
\newblock Towards accurate generative models of video: A new metric \& challenges, 2019.

\bibitem[Vaswani et~al.(2017)Vaswani, Shazeer, Parmar, Uszkoreit, Jones, Gomez, Kaiser, and Polosukhin]{vaswani2017attention}
Ashish Vaswani, Noam Shazeer, Niki Parmar, Jakob Uszkoreit, Llion Jones, Aidan~N Gomez, {\L}ukasz Kaiser, and Illia Polosukhin.
\newblock Attention is all you need.
\newblock \emph{NeurIPS}, 30, 2017.

\bibitem[{Vidu}(2024)]{vidu}
{Vidu}.
\newblock Vidu.
\newblock \url{https://www.vidu.com/}, 2024.

\bibitem[Wallace et~al.(2024)Wallace, Dang, Rafailov, Zhou, Lou, Purushwalkam, Ermon, Xiong, Joty, and Naik]{wallace2024diffusion}
Bram Wallace, Meihua Dang, Rafael Rafailov, Linqi Zhou, Aaron Lou, Senthil Purushwalkam, Stefano Ermon, Caiming Xiong, Shafiq Joty, and Nikhil Naik.
\newblock Diffusion model alignment using direct preference optimization.
\newblock In \emph{CVPR}, pages 8228--8238, 2024.

\bibitem[Wang et~al.(2016)Wang, Gan, Hu, Lin, Jin, Song, Wang, Katsavounidis, Aaron, and Kuo]{wang2016mcl}
Haiqiang Wang, Weihao Gan, Sudeng Hu, Joe~Yuchieh Lin, Lina Jin, Longguang Song, Ping Wang, Ioannis Katsavounidis, Anne Aaron, and C-C~Jay Kuo.
\newblock Mcl-jcv: a jnd-based h. 264/avc video quality assessment dataset.
\newblock In \emph{ICIP}, 2016.

\bibitem[Wang et~al.(2025)Wang, Lin, Wei, Zhao, Yang, Loy, and Jiang]{wang2025seedvr}
Jianyi Wang, Zhijie Lin, Meng Wei, Yang Zhao, Ceyuan Yang, Chen~Change Loy, and Lu~Jiang.
\newblock Seedvr: Seeding infinity in diffusion transformer towards generic video restoration.
\newblock \emph{arXiv preprint arXiv:2501.01320}, 2025.

\bibitem[Wang et~al.(2024)Wang, Bai, Tan, Wang, Fan, Bai, Chen, Liu, Wang, Ge, et~al.]{wang2024qwen2}
Peng Wang, Shuai Bai, Sinan Tan, Shijie Wang, Zhihao Fan, Jinze Bai, Keqin Chen, Xuejing Liu, Jialin Wang, Wenbin Ge, et~al.
\newblock Qwen2-vl: Enhancing vision-language model's perception of the world at any resolution.
\newblock \emph{arXiv preprint arXiv:2409.12191}, 2024.

\bibitem[Wang et~al.(2004)Wang, Bovik, Sheikh, and Simoncelli]{DBLP:journals/tip/WangBSS04}
Zhou Wang, Alan~C. Bovik, Hamid~R. Sheikh, and Eero~P. Simoncelli.
\newblock Image quality assessment: from error visibility to structural similarity.
\newblock \emph{IEEE TIP}, 13\penalty0 (4):\penalty0 600--612, 2004.
\newblock \doi{10.1109/TIP.2003.819861}.

\bibitem[Wu and He(2018)]{wu2018groupnormalization}
Yuxin Wu and Kaiming He.
\newblock Group normalization, 2018.

\bibitem[Xiao et~al.(2025)Xiao, Cheng, Qi, Gui, Cen, Ma, Yuille, and Jiang]{xiao2025videoauteur}
Junfei Xiao, Feng Cheng, Lu~Qi, Liangke Gui, Jiepeng Cen, Zhibei Ma, Alan Yuille, and Lu~Jiang.
\newblock Videoauteur: Towards long narrative video generation.
\newblock \emph{arXiv preprint arXiv:2501.06173}, 2025.

\bibitem[Yang et~al.(2024)Yang, Teng, Zheng, Ding, Huang, Xu, Yang, Hong, Zhang, Feng, Yin, Gu, Zhang, Wang, Cheng, Liu, Xu, Dong, and Tang]{yang2024cogvideoxtexttovideodiffusionmodels}
Zhuoyi Yang, Jiayan Teng, Wendi Zheng, Ming Ding, Shiyu Huang, Jiazheng Xu, Yuanming Yang, Wenyi Hong, Xiaohan Zhang, Guanyu Feng, Da~Yin, Xiaotao Gu, Yuxuan Zhang, Weihan Wang, Yean Cheng, Ting Liu, Bin Xu, Yuxiao Dong, and Jie Tang.
\newblock Cogvideox: Text-to-video diffusion models with an expert transformer, 2024.

\bibitem[Yin et~al.(2024)Yin, Zhao, Zheng, Lin, Ou, Chen, Huang, Wang, Tao, Wan, et~al.]{yin2024towards}
Yuanyang Yin, Yaqi Zhao, Mingwu Zheng, Ke~Lin, Jiarong Ou, Rui Chen, Victor Shea-Jay Huang, Jiahao Wang, Xin Tao, Pengfei Wan, et~al.
\newblock Towards precise scaling laws for video diffusion transformers.
\newblock \emph{arXiv preprint arXiv:2411.17470}, 2024.

\bibitem[Yu et~al.(2023{\natexlab{a}})Yu, Cheng, Sohn, Lezama, Zhang, Chang, Hauptmann, Yang, Hao, Essa, et~al.]{yu2023magvit}
Lijun Yu, Yong Cheng, Kihyuk Sohn, Jos{\'e} Lezama, Han Zhang, Huiwen Chang, Alexander~G Hauptmann, Ming-Hsuan Yang, Yuan Hao, Irfan Essa, et~al.
\newblock Magvit: Masked generative video transformer.
\newblock In \emph{CVPR}, 2023{\natexlab{a}}.

\bibitem[Yu et~al.(2023{\natexlab{b}})Yu, Lezama, Gundavarapu, Versari, Sohn, Minnen, Cheng, Birodkar, Gupta, Gu, et~al.]{yu2023language}
Lijun Yu, Jos{\'e} Lezama, Nitesh~B Gundavarapu, Luca Versari, Kihyuk Sohn, David Minnen, Yong Cheng, Vighnesh Birodkar, Agrim Gupta, Xiuye Gu, et~al.
\newblock Language model beats diffusion--tokenizer is key to visual generation.
\newblock \emph{arXiv preprint arXiv:2310.05737}, 2023{\natexlab{b}}.

\bibitem[Zhang et~al.(2018)Zhang, Isola, Efros, Shechtman, and Wang]{zhang2018unreasonableeffectivenessdeepfeatures}
Richard Zhang, Phillip Isola, Alexei~A. Efros, Eli Shechtman, and Oliver Wang.
\newblock The unreasonable effectiveness of deep features as a perceptual metric, 2018.

\bibitem[Zhao et~al.(2025)Zhao, Ni, Wang, Cheng, Yang, Jiang, and Wang]{zhao2025synthetic}
Qi~Zhao, Xingyu Ni, Ziyu Wang, Feng Cheng, Ziyan Yang, Lu~Jiang, and Bohan Wang.
\newblock Synthetic video enhances physical fidelity in video synthesis.
\newblock \emph{arXiv preprint arXiv:2503.20822}, 2025.

\bibitem[Zhao et~al.(2024)Zhao, Zhang, Cun, Yang, Niu, Li, Hu, and Shan]{zhao2024cvvaecompatiblevideovae}
Sijie Zhao, Yong Zhang, Xiaodong Cun, Shaoshu Yang, Muyao Niu, Xiaoyu Li, Wenbo Hu, and Ying Shan.
\newblock Cv-vae: A compatible video vae for latent generative video models, 2024.

\bibitem[Zhao et~al.(2023)Zhao, Gu, Varma, Luo, Huang, Xu, Wright, Shojanazeri, Ott, Shleifer, et~al.]{zhao2023pytorch}
Yanli Zhao, Andrew Gu, Rohan Varma, Liang Luo, Chien-Chin Huang, Min Xu, Less Wright, Hamid Shojanazeri, Myle Ott, Sam Shleifer, et~al.
\newblock Pytorch fsdp: experiences on scaling fully sharded data parallel.
\newblock \emph{arXiv preprint arXiv:2304.11277}, 2023.

\bibitem[Zheng et~al.(2024)Zheng, Peng, Yang, Shen, Li, Liu, Zhou, Li, and You]{opensora}
Zangwei Zheng, Xiangyu Peng, Tianji Yang, Chenhui Shen, Shenggui Li, Hongxin Liu, Yukun Zhou, Tianyi Li, and Yang You.
\newblock Open-sora: Democratizing efficient video production for all, March 2024.

\end{thebibliography}
